\documentclass[12pt]{article}
\usepackage[margin=1in]{geometry}
\usepackage{setspace}
\usepackage{lmodern}
\usepackage[utf8]{inputenc} 
\usepackage[T1]{fontenc} 
\usepackage{hyperref}       
\usepackage{url}            
\usepackage{booktabs}       
\usepackage{amsfonts}       
\usepackage{nicefrac}       
\usepackage{microtype}      
\usepackage{comment}
\usepackage[leqno]{amsmath}
\usepackage{epsfig,latexsym, amsthm}
\usepackage{mathtools}
\usepackage{float}
\usepackage{subfig}
\usepackage[font={small}]{caption}
\usepackage{algorithm, algpseudocode}
\usepackage{afterpage}

\newtheorem{theorem}{Theorem}
\newtheorem{lemma}{Lemma}

\newtheorem{assumption}{Assumption}

\newtheorem{remark}{Remark}
\newtheorem{definition}{Definition}
\newcommand{\field}[1]{\mathbb{#1}}

\def\A{\mathcal{A}}
\def\N{\mathcal{N}}
\def\W{\mathcal{A}}
\def\w{a}                    
\def\x{x}  

\def\p{\mathcal{P}}
         
\def\V{\field{V}}                                
\def\X{\mathcal{X}}

\def\R{\field{R}}                                
\def\E{\mathbb{E}}
\def\V{\mathbb{V}}

\DeclareMathOperator*{\argmin}{argmin}
\DeclareMathOperator*{\argmax}{argmax}

\newcounter{proofstep}
\newenvironment{proofstep}[1]
{\vspace{3pt}
	\refstepcounter{proofstep}%
	\par\textit{Step~\arabic{proofstep}:~#1}.\,}
{\vspace{2pt}}

\def\B{B}

\def\I{\mathbf{I}}
\title{Estimation Considerations in Contextual Bandits}

\author{
{\fontsize{12}{12} \selectfont Maria Dimakopoulou}\thanks{Stanford University, Management Science and Engineering, madima@stanford.edu} \and
{\fontsize{12}{12} \selectfont Zhengyuan Zhou} \thanks{Stanford University, Electrical Engineering, zyzhou@stanford.edu} \and
{\fontsize{12}{12} \selectfont Susan Athey} \thanks{Stanford University, Graduate School of Business, and NBER, athey@stanford.edu} \and
{\fontsize{12}{12} \selectfont Guido Imbens} \thanks{Stanford University, Graduate School of Business, and NBER, imbens@stanford.edu}
}

\date{}
\begin{document}
\maketitle

\begin{abstract}
Contextual bandit algorithms are sensitive to the estimation method of the outcome model as well as the exploration method used, particularly in the presence of rich heterogeneity or complex outcome models, which can lead to difficult estimation problems along the path of learning.
We study a consideration for the exploration vs. exploitation framework that does not arise in multi-armed bandits but is crucial in contextual bandits; the way exploration and exploitation is conducted in the present affects the bias and variance in the potential outcome model estimation in subsequent stages of learning. 
We develop parametric and non-parametric contextual bandits that integrate balancing methods from the causal inference literature in their estimation to make it less prone to problems of estimation bias. We provide the first regret bound analyses for contextual bandits with balancing in the domain of linear contextual bandits that match the state of the art regret bounds. We demonstrate the strong practical advantage of balanced contextual bandits on a large number of supervised learning datasets and on a synthetic example that simulates model mis-specification and prejudice in the initial training data.
Additionally, we develop contextual bandits with simpler assignment policies by leveraging sparse model estimation methods from the econometrics literature and demonstrate empirically that in the early stages they can improve the rate of learning and decrease regret.
\end{abstract}

\newpage

\section{Introduction}\label{sec:intro}
Contextual bandits seek to learn a personalized treatment assignment policy in the presence of treatment effects that vary with observed contextual features.
In such settings, there is a need to balance the exploration of treatments\footnote{A treatment is also referred to as an arm in the literature. In this paper, we use the two terms interchangeably.} for which there is limited knowledge in order to improve performance in the future against the exploitation of existing knowledge in order to attain better performance in the present (see \cite{bubeck-survey} for a survey). 
Since large amounts of data can be required to learn how the benefits of alternative treatments vary with individual characteristics, contextual bandits can play an important role in making experimentation and learning more efficient.
Several successful contextual bandit designs have been proposed \cite{auer-linrel, li-linucb, agrawal-lints, agarwal-ilovetoconbandits,bastani2015online}. 
The existing literature has provided regret bounds (e.g., the general bounds of \cite{russo-vanroy}, the bounds of  \cite{rigollet-nonparamtheory, perchet-nonparamtheory, slivkins-nonparamtheory} in the case of non-parametric function of arm rewards), has demonstrated successful applications (e.g., news article recommendations \cite{li-linucb} or mobile health \cite{lei-mhealth}), and has proposed system designs to apply these algorithms in practice \cite{agarwal-debt}.

Contextual bandits are poised to play an important role in a wide range of applications:
content recommendation in web-services, where the learner wants to personalize recommendations (arm) to the profile of a user (context) to maximize engagement (reward);  online education platforms, where the learner wants to select a teaching method (arm) based on the characteristics of a student (context) in order to maximize the student's scores (reward); and survey experiments, where the learner wants to learn what information or persuasion (arm) influences the responses (reward) of subjects as a function of their demographics, political beliefs, or other characteristics (context).

In the contextual setting, one does not expect to see many future observations with the same context as the current observation, and so the value of learning from pulling an arm for this context accrues when that observation is used to estimate the outcome from this arm for a different context in the future.
Therefore, the performance of contextual bandit algorithms can be sensitive to the estimation method of the outcome model or the exploration method used.  
In the initial phases of learning when samples are small, biases are likely to arise in estimating the outcome model using data from previous non-uniform assignments of contexts to arms.
The bias issue is aggravated in the case of a mismatch between the generative model and the functional form used for estimation of the outcome model, or similarly, when the heterogeneity in treatment effects is too complex to estimate well with small datasets.  
In that case methods that proceed under the assumption that the functional form for the outcome model is correct may  be overly optimistic about the extent  of the learning so far, and emphasize exploitation over exploration.
Another case where biases can arise occurs when training observations from certain regions of the context space are scarce (e.g., prejudice in training data if a non-representative set of users arrives in initial batches of data). 
These problems are common in real-world settings, such as in survey experiments in the domain of social sciences or in applications to health, recommender systems, or education.  
For example, early adopters of an online course may have different characteristics than later adopters.  

We develop parametric and non-parametric contextual bandits that integrate balancing methods from the causal inference literature \cite{imbens-ci} in their estimation to make it less prone to the aforementioned sources of bias.
Our methods aim to balance covariates between treatment groups and achieve contextual bandit designs which are less prone to problems of bias.
The balancing will lead to lower estimated precision in the reward functions, and thus will emphasize exploration longer than the conventional linear TS and UCB algorithms, leading to more robust estimates.
Balancing can take various forms, ranging from the well-known inverse propensity score weighting to state of the art methods such as approximate residual balancing \cite{athey-arb} or the method of \cite{kallus-offline}.
Moreover, balancing can be integrated in contextual bandits with a parametric model estimation such as ridge or LASSO) or non-parametric model estimation such as forests.

We further investigate the effect of balancing via inverse propensity weighting in the domain of linear contextual bandits, by comparing linear Thompson sampling (LinTS) \cite{agrawal-lints} and linear upper confidence bound (LinUCB) \cite{li-linucb} -- which have strong theoretical guarantees -- with our algorithms, \textit{balanced linear Thompson sampling} (BLTS) and \textit{balanced linear upper confidence bound} (BLUCB).
Our main contribution here is to provide extensive and convincing empirical evidence for the effectiveness of BLTS and BLUCB (in comparison to LinTS and LinUCB) by considering the problem of multiclass classification with bandit feedback. 
Specifically, we transform a $K$-class classification task into a $K$-armed contextual bandit \cite{dudik-offline-1} and we use 300 public benchmark datasets for our evaluation.
Additionally, we provide regret bounds for BLTS and BLUCB, which match the existing state-of-the-art regret bounds for LinTS and LinUCB.
It is important to point out that, even though BLTS and LinTS share the same theoretical guarantee, BLTS outperforms LinTS empirically. Similarly, BLUCB has a strong empirical advantage over LinUCB.
In bandits, this phenomenon is not uncommon. For instance, it is well-known that even though the existing UCB bounds are often tighter than those of Thompson sampling, Thompson sampling performs better in practice than UCB~\cite{chapelle-tsucb}. Consequently, we take the view that even though regret is a useful theoretical performance metric, it may not always provide clear guidance on which algorithm should be used in practice.
We find that this is also the case for balanced linear contextual bandits, as in our evaluation BLTS has a strong empirical advantage over BLUCB. 
Overall, in this large-scale evaluation, BLTS outperforms LinUCB, BLUCB and LinTS.
In our empirical evaluation, we also consider a synthetic example  which simulates in the simplest possible way two issues of bias that often arise in practice; training data with non-representative contexts and model mis-specification.
BLTS is much more effective in escaping these biases and, as in the evaluation on supervised learning datasets,
it outperforms LinUCB, BLUCB and LinTS by a large margin.

To our knowledge, this is the first work to integrate balancing in the online contextual bandit setting, to perform a large-scale evaluation of it against direct estimation method baselines with theoretical guarantees and to provide a theoretical characterization of balanced contextual bandits that match the regret bound of their direct method counterparts.
The balancing technique is well-known in machine learning, especially in domain adaptation and  studies in learning-theoretic frameworks \cite{huang-ml,zadrozny-ml,cortes-ml}.
There is a number of recent works which approach contextual bandits through the framework of causality \cite{bareinboim-bandits, bareinboim-fusion, forney-fusion, lattimore-causalbandit}. 
There is also a significant body of research that leverages balancing for offline evaluation and learning of contextual bandit or reinforcement learning policies from logged data \cite{strehl-offline, dudik-offline-1, li-offline-1, dudik-offline-2, li-offline-2, swaminathan-offline, jiang-offline, thomas-offline, athey-offline, kallus-offline, wang-offline, deshpande-offline, kallus2018policy, strehl2010learning,zhou2018offline}. 
In the offline setting, the complexity of the historical assignment policy is taken as given, and thus the difficulty of the offline evaluation and learning of optimal policies is taken as given. 
Therefore, these results lie at the opposite end of the spectrum from our work, which focuses on the online setting, where the complexity of the assignment policy is known and controlled by the algorithm. 
Methods for reducing the bias due to adaptive data collection have also been studied for non-contextual multi-armed bandits \cite{villar-online, nie-online}, but the nature of the estimation in contextual bandits is qualitatively different. 
Importance weighted regression in contextual bandits was first mentioned in \cite{agarwal-ilovetoconbandits}, but without a systematic motivation, analysis and evaluation of this technique. To our knowledge, our paper is the first work to integrate balancing in the online contextual bandit setting, to perform a large-scale evaluation of it against direct estimation method baselines with theoretical guarantees and to provide a theoretical characterization of balanced contextual bandits that match the regret bound of their direct method counterparts. The effect of importance weighted regression is also evaluated in \cite{bietti2018contextual}, but this paper is a successor to the extended version of our paper.

Reweighting or balancing methods address model misspecification by making the estimation ``doubly-robust,'', robust against misspecification of the reward function, important here, and robust against the specification of the propensity score (not as important here because in the bandit setting we know the propensity score).  The term ``doubly-robust'' comes from the extensive literature on offline policy evaluation \cite{scharfstein1999adjusting}; it means that when comparing two policies using historical data, we get consistent estimates of the average difference in outcomes for segments of the context whether we have either a well-specified model of rewards  or not., because  we   have good model of the arm assignment policy (i.e., accurate propensity scores). In a contextual bandit, the learner controls the arm assignment policy conditional on the observed context and therefore has access to accurate propensity scores even in small samples. So, even when the reward model is severely misspecified,  the learner can obtain more accurate value estimates of the reward function for each value of the context.

In real-world applications, such as health, education, recommender systems, there may be contextual variables that are highly predictive of user outcomes (e.g. previous health metrics, previous test scores, or previous consumer choices), but less important for optimizing arm assignment. We continue by showing that such contextual variables can be particularly problematic in terms of generating bias and variance in the early stages of learning.
Even though in principle a variety of methods can
be used to consistently estimate outcome models in the presence of complex, non-uniform arm assignment in large samples (Ridge regression as in \cite{agrawal-lints, li-linucb}, ordinary least squares as in \cite{goldenshluger-olsucb}, LASSO as in \cite{bastani-lassobandit}), in the early stages of learning where samples are small, these methods are only partially effective, and so the extent to which the method exacerbates the estimation problem
will affect performance.  
In the domain of linear contextual bandits, we simplify the assignment policy by simplifying the estimated outcome model. 
We develop the bootstrap LASSO Thompson sampling and UCB contextual bandits, which use $L_1$ regularization in the estimation of the outcome models and circumvent the lack of closed-form solution by using the bootstrap to form a sampling distribution of the outcomes in order to obtain an approximate posterior for Thompson sampling and an upper confidence bound for UCB.
We also propose a method to simplify the assignment policy with a form of smoothing that partitions the context space via a classification tree and defines the assignment rule for each partition rather than for each context distinctly.
We show that, all else equal, using assignment policies that are simpler (in terms of how they vary with contextual variables) in the early learning phases of the algorithm can improve the rate of learning and decrease regret.
Simple assignment rules may have other advantages as well; for example, \cite{lei-mhealth} highlights the advantages of simplicity for interpretability in health applications of contextual bandits.

\section{Methodological Designs in Contextual Bandits}\label{sec:designs}

In contextual bandits,
 contexts $\x_1, \x_2, \dots, \x_T \in \mathcal{X} \subset \R^d$ arrive sequentially.
We have a finite set of $K$ arms, $\W = \{1,2,\dots, K\}$, which we wish to assign to each context upon its arrival. 
We posit that the data $(\x_t, r_t(1), r_t(2), \dots, r_t(K))$ are drawn \textbf{iid} from a fixed underlying joint distribution on $(\x, r(1), r(2), \dots, r(K))$, where $r(\w) \in \mathbb{R}$ denotes the (random) reward under arm $\w$ and context $\x$. Note that the marginal distribution on $x$ specifies the distribution from which the contexts are drawn.
The observables are $(\x_t, \w_t, r_t(\w_t))$; in particular, only the reward $r_t(\w_t)$  for the chosen arm $\w_t$ is observed.
For each context $\x\in \R^d$, the optimal assignment is $\w^*(\x) = \argmax_{\w}\left\{\E[r(\w) | \x]\right\}$ and we let $\w_t^* = \w^*(\x_t)$, which denotes
the optimal assignment for context $x_t$.
The objective is to find an assignment rule that sequentially assigns $\w_t$ to minimize the cumulative expected regret $\sum_{t=1}^\top \E [r(\w^*_t) - r(\w_t)]$, where the assignment rule is a function of the previous observations $\left(\x_j, \w_j, r(\w_j)\right)$ for $j=1,\dots,t-1$ and of the new context $\x_t$.
We next discuss three methodological designs in contextual bandits.

\subsection{Model Estimation}\label{subsec:direct}
An important component for sequential arm assignment lies in modeling and estimating the conditional expected reward corresponding to each arm $a \in \A$ given context $x$, $\mu_\w(\x) =  \E[r(\w) |  \x]$. 
In a contextual bandit there are as many models to be estimated as arms.
We do this estimation separately for each arm $\w \in \W$ on the history of observations corresponding to this arm $\{\left(\x_t, \w_t, r_t(\w_t)\right) \mid \w_t = \w\}$. 

\subsubsection{Parametric Estimation} \label{bootstrap}
In parametric estimation, we estimate  $\mu_\w(\x)$ by a parametric form. A commonly used model in contextual bandits is the linear model,
where $\mu_\w(\x) =  \x^\top \theta_\w$, with unknown parameters $\theta_\w$.
More generally, estimation can be done via generalized linear models\footnote{GLM bandits are discussed in \cite{li-glmucb}.} $\mu_\w(\x) =  g(\x^\top \theta_\w)$, where $g: \R \to \R$ is a strictly increasing and known link function. 
Denote \textbf{r}$_a$ as the response vector and \textbf{X}$_a$ as the covariate matrix of the history of observations assigned to $\w$. 
The model parameters can be estimated via $L_1$ (LASSO \cite{tibshirani-lasso} in linear models) or $L_2$  (ridge \cite{hoerl-ridge} in linear models) regularized regression. 
For a new context $x$, we wish to obtain the conditional mean $\hat{\mu}_\w(\x)$ of the reward associated with each arm $a \in A$ and its variance $\V(\hat{\mu}_a(x))$.
In some cases, the estimates can be computed in closed form (e.g. in the case of a linear model estimated with ridge regression). 
However, in many cases (e.g. in the case of a linear model estimated with LASSO regression, or generalized linear models) exact computation is intractable and we must perform approximation \cite{russo2018tutorial}.
When exact computation is intractable, bootstraping provides a viable way to approximate these quantities. More specifically, we can obtain a sampling distribution on $\mu_a(x)$ by training many regularized regression models on bootstrap samples drawn from $(\textbf{X}_a, \textbf{r}_a)$. With this sampling distribution, we can then easily to compute the mean estimate $\hat{\mu}_\w(\x)$ and the variance estimate $\V(\hat{\mu}_a(x))$. Note that as long as one can solve the underlying regression problem efficiently (either in closed form or via some fast numeric scheme), the estimates can be constructed by bootstraping.
Consequently, this provides a general-purpose way of computing $\hat{\mu}_\w(\x)$ and estimate $\V(\hat{\mu}_a(x))$ for contextual bandits. 

\subsubsection{Non-parametric Estimation} \label{grf}
Parametric estimation can have high bias when the model is mis-specified (also called unrealizable in the literature). Non-parametric models, on the other hand, are more expressive and comparatively suffer less from the bias problem. In this paper, we consider non-parametric estimation of $\mu_\w(\x)$ by training a generalized random forest \cite{athey-grf} on the history of observations assigned to $\w$.
Generalized random forest is an ensemble method and hence provides estimates of the conditional mean $\hat{\mu}_\w(\x)$ and variance $\V(\hat{\mu}_a(x))$. 
Related approaches based on random forests or decision trees have been proposed in \cite{feraud-banditforest, elmachtoub-decisiontreesbootstrap}. 
The generalized random forest is a method that preserves the core elements of random forests \cite{breiman-rf} including recursive partitioning, sub-sampling, and random split selection, but differs from traditional approaches in several ways, most notably that it integrates ``honest'' tree estimation \cite{athey-ct}: the sample used to select the splits of the tree is independent from the sample used to estimate the improvement in fit yielded by a split.  
The ``honesty'' property of generalized random forests reduces bias and overfitting to outliers, which may be of particular concern in early stages of learning.  In cases where the outcome functional form is complicated, as is often the case in practice, bandits based on non-parametric estimation tend to perform better. In addition, they can flexibly control for features which are confounders for the estimation of the reward functions (i.e., features that affect outcomes and were also used to determine assignments in earlier contexts).

\subsection{Treatment Assignment Rules}\label{subsec:rules}
Thompson sampling \cite{thompson-ts, scott-ts, agrawal-ts, russo-tstutorial} and upper confidence bound (UCB) \cite{lai-ucb, auer-ucb} are two different methods for assigning contexts to arms which are highly effective in dealing with the exploration-exploitation trade-off. 
In both methods, until every arm has been pulled at least once, the first contexts are assigned to arms in $\W$ at random with equal probability. 
At every time $t$ and for every arm $\w$, the two methods use the history of observations $\{\left(\x_t, \w_t, r_t(\w_t)\right) \mid \w_t = \w\}$, to obtain the estimates of the functions $\hat{\mu}_\w(\x)$ and $\V(\hat{\mu}_a(x))$ with the methods outlined in Section \ref{subsec:direct}.

Thompson sampling assumes that the expected reward $\mu_a(x_t)$ associated with arm $a$ conditional on the context $x_t$ is Gaussian $ \N\left(\hat{\mu}_a(x_t), \alpha^2\V(\hat{\mu}_a(x_t)) \right)$, where $\alpha$ is an appropriately chosen constant.
Then, it draws a sample $\tilde{\mu}_a(x_t)$ from the distribution of each arm $a \in \cal$ and context $x_t$ is then assigned to the arm with the highest sample, $a_t = \argmax_a \{\tilde{\mu}_{a}(x_t)\}$.

On the other hand, UCB computes upper confidence bounds for the expected reward $\mu_a(x_t)$ of context $x_t$ associated with each arm $a \in \A$ and assigns the context to the arm with the highest upper confidence bound, $a_t = \argmax_a\left\{\hat{\mu}_a(x_t) + \alpha \sqrt{\V(\hat{\mu}_a(x_t))}\right\}$, where $\alpha$ is an appropriately chosen constant.

\subsection{Balancing in Contextual Bandits}\label{subsec:ipw}

In this section, we use balancing methods from the causal inference literature to improve the existing UCB and Thompson sampling algorithms by balancing features between arms to reduce bias. 
We focus on the method of inverse propensity weighting (IPW) \cite{imbens-ci}.
Denote \textbf{r} as the reward vector, \textbf{X} as the context matrix and \textbf{a} as the arm assignment vector of all previous observations. 

For UCB, we train a multi-class logistic regression model of \textbf{a} on \textbf{X} to estimate the assignment probabilities $p_\w(\x), \w \in \W$, also known as propensity scores.
Note that UCB has deterministic assignment rules, so that conditional on the batch, the propensity scores are either zero or one. However, we can consider the ordering of contexts' arrival as random and use $\hat{p}_\w(\x)$ as a balancing weight to account for non-uniform assignment in previous contexts. 

For Thompson sampling, the propensity scores are in principle known because Thompson sampling performs probability matching, i.e., it assigns a context to an arm with the probability that this arm is optimal.
Since computing the propensity scores involves high order integration, they can be approximated via Monte-Carlo simulation. 
Each iteration draws a sample from the posterior reward distribution of every arm $a$ conditional on $x$, where the posterior is the one that the algorithm considered at the end of a randomly selected prior time period.  
The propensity score $p_a(x)$ is the fraction of the Monte-Carlo iterations in which arm $a$ has the highest sampled reward, where the arrival time of context $x$ is treated as random.


With these propensity estimates, we can modify the model estimation as follows.
For parametric estimation, we define the weight of each observation $(x, a, r)$ as the inverse of the estimated propensity score, 
$$w_a = 1/\hat{p}_{\w}(\x)$$ and we train the weighted counterparts of the regularized regressions discussed in Section \ref{bootstrap}. 
For non-parametric estimation (via generalized random forest), one alternative is to construct an augmented covariate matrix $\tilde{\textbf{X}}_a$ and reward vector $\tilde{\textbf{r}}_a$ by replicating $[w_a]$ times each observation $(x,a,r)$ and subsequently to estimate the generalized random forest on $\tilde{\textbf{X}}_a$ and $\tilde{\textbf{r}}_a$. 
Another alternative is to treat the propensity score of each observation as a contextual variable and estimate the generalized random forest on  $[\textbf{X}_a : \textbf{p}_a]$ and $\textbf{r}_a$, where $\textbf{p}_a$ is the vector of the propensity scores for previous observations for arm $a$.

In both cases, weighting the observations by the inverse propensity scores reduces bias, but even when the propensity scores are known it increases variance, particularly when they are small. 
Consequently, since eventually assignment probabilities should approach zero or one for all arms and contexts clipping the propensity scores \cite{crump2009dealing,kallus2018policy} with some threshold  $\gamma$, e.g. $0.1$ helps control the variance increase. 

Finally, note that one could integrate in the contextual bandit estimation other covariate balancing methods, such as the method of approximate residual balancing \cite{athey-arb} or the method of \cite{kallus-offline}. 
For instance, with approximate residual balancing one would use as weights
$$w_a = \argmin_w \left\{ (1- \zeta) \lVert w\rVert_2^2 + \zeta \lVert \bar{x} - \textbf{X}_a^\top w \rVert_\infty^2 \text{ s.t. } \sum_{t: a_t = a} w_t = 1 \text{ and } 0 \leq w_t \leq n_a^{-2/3} \right\}$$
where $\zeta \in (0, 1)$ is a tuning parameter, $n_a =  \sum_{t = 1}^{T} \textbf{1} \{a_t = a\}$ and $\bar{x} = \frac{1}{T} \sum_{t = 1}^{T} x_t$ and then use $w_a$ to modify the parametric and non-parametric model estimation as outlined before. 


 \section{The Effects of Balancing}\label{sec:ucbvsts}
In this section, we study empirically and theoretically the effects of balancing as introduced in~\ref{subsec:ipw}. For concreteness, we focus on the method of inverse propensity weighting for balancing and on linear models with $L_2$ regularization (i.e. ridge) and refer to the resulting algorithms as \textit{balanced linear Thompson sampling} (BLTS) and \textit{balanced linear upper confidence bound} (BLUCB), as given in Algorithm~\ref{alg:BLTS} and Algorithm~\ref{alg:BLUCB}. 

BLTS and BLUCB build on linear contextual bandits LinTS \cite{agrawal-lints} and LinUCB \cite{li-linucb} respectively. 
In LinTS and LinUCB, the expected reward is assumed to be a linear function of the context $x_t$ with some unknown coefficient vector $\theta_a$, $\E[r_t(a) | x_t = x] = x^\top \theta_a$, and the variance is typically assumed to be constant $\V[r_t(a) | x_t = x] = \sigma^2_a$.
At time $t$, LinTS and LinUCB apply ridge regression with regularization parameter $\lambda$ to the history of observations $(X_a, r_a)$ for each arm $a \in \A$, in order to obtain an estimate $\hat{\theta}_a$ and its variance $\V_a(\hat\theta_a)$.
For the new context $x_t$, $\hat{\theta}_a$ and its variance are used by LinTS and LinUCB to obtain the conditional mean $\hat{\mu}_a(x_t)=x_t^\top\hat\theta_a$  of the reward associated with each arm $a \in A$, and its  variance $\V(\hat{\mu}_a(x_t))=x^\top_t \V(\hat\theta_a) x_t $.

BLTS and BLUCB are linear contextual bandit algorithms that perform balanced estimation of the model of all arms in order to obtain a Gaussian distribution and an upper confidence bound respectively for the reward associated with each arm conditional on the context.
The idea is that at every time $t$, the linear contextual bandit weighs each observation $(x_\tau, a_\tau, r_\tau(a_\tau))$, $\tau = 1, \dots, t$ in the history up to time $t$ by the inverse propensity score, $p_{a_\tau}(x_\tau)$. Then, for each arm $a \in \A$, the linear contextual bandit weighs each observation $(x_\tau, a, r_\tau(a))$ in the history of arm $a$ by $w_{a} = 1 / p_{a}(x_\tau)$ and uses weighted regression to obtain the estimate $\hat{\theta}^\text{balanced}_a$ with variance $\V(\hat\theta^\text{balanced}_a)$.

In BLTS (Algorithm~\ref{alg:BLTS}), the observations are weighted by the known propensity scores.
For every arm $a$, the history $(X_a, r_a, p_a)$ is used to obtain a balanced estimate $\hat{\theta}^\text{BLTS}_a$ of $\theta_a$ and its variance $\V(\hat\theta^\text{BLTS}_a)$ which produce a normally distributed estimate of $\tilde{\mu}_a \sim \N\left(x_t^\top \hat{\theta}^\text{BLTS}_a, \alpha^2 x_t^\top \V(\hat\theta^\text{BLTS}_a) x_t\right)$ of the reward of arm $a$ for context $x_t$, where $\alpha$ is a parameter of the algorithm.

In BLUCB (Algorithm~\ref{alg:BLUCB}), the observations are weighted by the estimated propensity scores.
For every arm $a$, the history $(X_a, r_a, \hat{p}_a)$ is used to obtain a balanced estimate $\hat{\theta}^{\text{BLUCB}}_a$ of $\theta_a$ and its variance $\V(\hat\theta^\text{BLUCB}_a)$. These are used to construct the upper confidence bound, $x_t^\top \hat{\theta}_a + \alpha \sqrt{x_t^\top \V(\hat\theta^\text{BLUCB}_a) x_t}$, for the reward of arm $a$ for context $x_t$,  where $\alpha$ is a constant. (For some results, e.g., \cite{auer2002using},  $\alpha$ needs to be slowly increasing in $t$.)

\begin{algorithm}[t]
	\caption{Balanced Linear Thompson Sampling} 
	\label{alg:BLTS}
	\begin{algorithmic}[1]
		\State \textbf{Input:} Regularization parameter $\lambda > 0$, propensity score threshold $\gamma \in (0, 1)$, constant $\alpha$ (deafult is 1)
		\State Set $\hat{\theta}^{\text{BLTS}}_a \leftarrow \textbf{null}, \B_a \leftarrow \textbf{null}, \forall a \in \mathcal{A}$
		\State Set $X_a \leftarrow$ empty matrix, $r_a \leftarrow$ empty vector $\forall a \in \mathcal{A}$
		\For {$t = 1, 2, \dots, T$}
		\If {$\exists a \in \A \text{ s.t. }\hat{\theta}^{\text{BLTS}}_a = \textbf{null}$ or $\B_a = \textbf{null}$}
		\State Select $a \sim \text{Uniform}(\A)$
		\Else
		\State Draw $\tilde{\theta}_a$ from $\N\left(\hat{\theta}^{\text{BLTS}}_a, \alpha^2 \V(\hat{\theta}^{\text{BLTS}}_a) \right)$ for all $a \in \A$ 
		\State Select $a = \arg\max\limits_{a\in\A} x_t^\top \tilde{\theta}_a$
		\EndIf
		\State Observe reward $r_t(a)$.
		\State Set $W_a \leftarrow $ empty matrix
		\For {$\tau = 1, \dots, t$}
		\If {$a_\tau = a$}
		\State Compute $p_a(x_\tau)$ and set $w = \frac{1}{\max(\gamma, p_a(x_\tau))}$
		\State $W_a \leftarrow \text{diag}(W_a, w)$
		\EndIf
		\EndFor
		\State $X_a \leftarrow [X_a : x_t^\top]$
		\State $B_a \leftarrow X_a^\top W_a X_a + \lambda \I$
		\State $r_a \leftarrow [r_a : r_t(a)]$ 
		\State $\hat{\theta}^{\text{BLTS}}_a \leftarrow B_a^{-1} X_a^\top W_a r_a$
		\State $\V(\hat{\theta}^{\text{BLTS}}_a)\leftarrow  B_a^{-1}\left((r_a-X_a^\top\hat{\theta}^{\text{BLTS}}_a)^\top W_a(r_a-X_a^\top\hat{\theta}^{\text{BLTS}}_a)\right) $
		\EndFor
	\end{algorithmic}
\end{algorithm}

\begin{algorithm}[t]
	\caption{Balanced Linear UCB} 
	\label{alg:BLUCB}
	\begin{algorithmic}[1]
		\State \textbf{Input:} Regularization parameter $\lambda > 0$, propensity score threshold $\gamma \in (0, 1)$, constant $\alpha$.
		\State Set $\hat{\theta}^{\text{BLUCB}}_a \leftarrow \textbf{null}, \B_a \leftarrow \textbf{null}, \forall a \in \mathcal{A}$
		\State Set $X_a \leftarrow$ empty matrix, $r_a \leftarrow$ empty vector $\forall a \in \mathcal{A}$
		\For {$t = 1, 2, \dots, T$}
		\If {$\exists a \in \A \text{ s.t. }\hat{\theta}^{\text{BLUCB}}_a = \textbf{null}$ or $\B_a = \textbf{null}$}
		\State Select $a \sim \text{Uniform}(\A)$
		\Else
		\State Select $a = \arg\max\limits_{a\in\A} \left(x_t^\top \hat{\theta}^{\text{BLUCB}}_a + \alpha \sqrt{x_t^\top  \V(\hat{\theta}^{\text{BLUCB}}_a) x_t}\right)$
		\EndIf
		\State Observe reward $r_t(a)$.
		\State Set $W_a \leftarrow $ empty matrix
		\For {$\tau = 1, \dots, t$}
		\If {$a_\tau = a$}
		\State Estimate $\hat{p}_a(x_\tau)$ and set $w = \frac{1}{\max(\gamma, \hat{p}_a(x_\tau))}$
		\State $W_a \leftarrow \text{diag}(W_a, w)$
		\EndIf
		\EndFor
		\State $X_a \leftarrow [X_a : x_t^\top]$
		\State $B_a \leftarrow X_a^\top W_a X_a + \lambda \I$
		\State $r_a \leftarrow [r_a : r_t(a)]$ 
		\State $\hat{\theta}_a \leftarrow B_a^{-1} X_a^\top W_a r_a$
		\State $\V(\hat{\theta}^{\text{BLUCB}}_a)\leftarrow  B_a^{-1}\left((r_a-X_a^\top\hat{\theta}^{\text{BLUCB}}_a)^\top W_a(r_a-X_a^\top\hat{\theta}^{\text{BLUCB}}_a)\right) $
		\EndFor
	\end{algorithmic}
\end{algorithm}

We present computational results that compare the performance of BLTS and BLUCB, with the existing state-of-the-art linear contextual bandits algorithms LinTS \cite{agrawal-lints} and LinUCB \cite{li-linucb}.
More specifically, we first present a simple synthetic example that simulates bias in the training data by under-representation or over-representation of certain regions of the context space and investigates the performance of the considered linear contextual bandits both when the outcome model of the arms matches the true reward generative process and when it does not match the true reward generative process.
Second, we conduct an experiment by leveraging 300 public, supervised cost-sensitive classification datasets to obtain contextual bandit problems, treating the features as the context, the labels as the arms and revealing only the reward for the chosen label.
We show that BLTS performs better than LinTS and that BLUCB performs better than LinUCB. 
The randomized assignment nature of Thompson sampling facilitates the estimation of the arms' outcomes models compared to UCB, and as a result LinTS outperforms LinUCB and BLTS outperforms BLUCB. Overall, BLTS has the best performance.
In the supplemental material, we include experiments against the policy-based contextual bandit from \cite{agarwal-ilovetoconbandits} which is statistically optimal but it is also outperformed by BLTS.
Finally, we give a theoretical guarantee that matches the
existing regret performance of BLTS and BLUCB. 

\subsection{A Synthetic Example}\label{subsec:pedagogic}
\begin{figure}[t]
	\centering
	\includegraphics[width=0.4\textwidth]{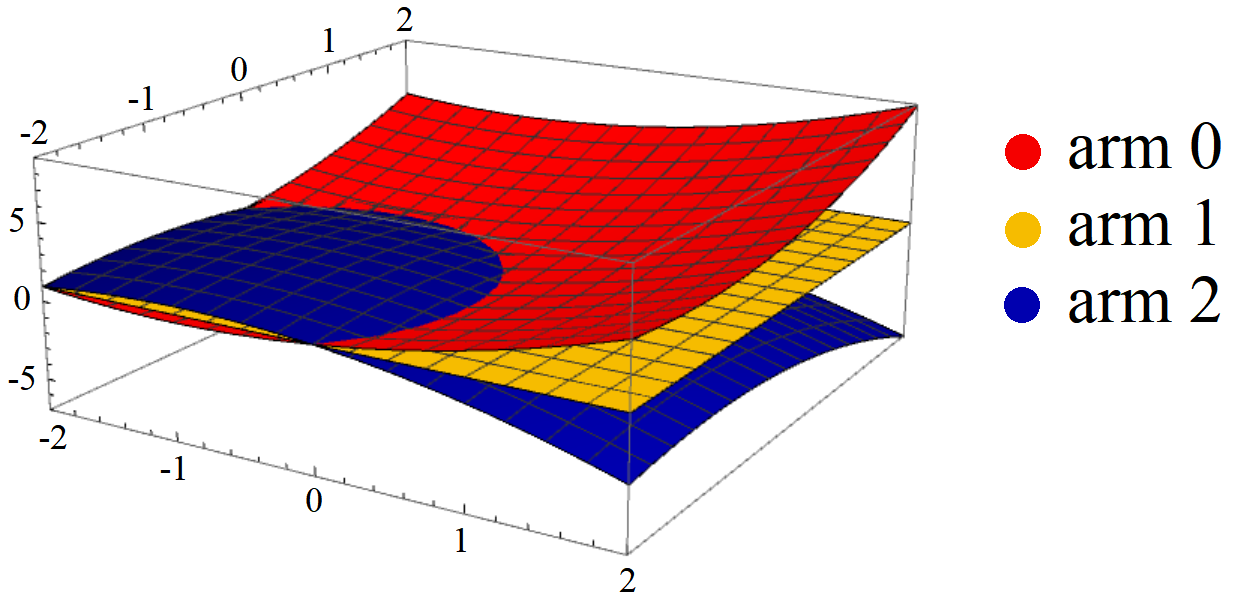}
	\caption{Expectation of each arm's reward, $\E[r_t(0)] = 0.5 (x_{t,0} + 1)^2 + 0.5 (x_{t,1} + 1)^2$ (red), $\E[r_t(1)] = 1$ (yellow), $\E[r_t(2)] = 2 - 0.5 (x_{t,0} + 1)^2 - 0.5 (x_{t,1} + 1)^2$ (blue).}
	\label{PotentialOutcomes}
\end{figure}

This simulated example aims to reflect in a simple way two issues that often arise in practice.
The first issue is the presence of bias in the training data by under-representation or over-representation of certain regions. A personalized policy that is trained based on such data and is applied to the entire context space will result in biased decisions for certain contexts.  The second issue is the problem of mismatch between the true reward generative process and the functional form used for estimation of the outcome model of the arms, which is common in applications with complex generative models. Model mis-specification aggravates the presence of bias in the learned policies.

We use this simple example to present in an intuitive manner why balancing and randomized assignment rule help with these issues, before moving on to a large-scale evaluation of the algorithms in real datasets in the next section.

Consider a simulation design where there is a warm-start batch of training observations, but it consists of contexts focused on one region of the context space. 
There are three arms $\A = \{0, 1, 2\}$ and the contexts $x_t = (x_{t,0}, x_{t,1})$ are two-dimensional with $x_{t,j} \sim \mathcal{N}(0, 1), \enskip j \in \{0, 1\}$. 
The rewards corresponding to each arm $a \in \A$ are generated as follows; $r_t(0) = 0.5 (x_{t,0} + 1)^2 + 0.5 (x_{t,1} + 1)^2 + \epsilon_t$, $r_t(1) = 1 + \epsilon_t$, and $r_t(2) = 2 - 0.5 (x_{t,0} + 1)^2 - 0.5 (x_{t,1} + 1)^2 +\epsilon_t$, where $\epsilon_t \sim \mathcal{N}(0, \sigma^2)$, $\sigma^2 = 0.01$. 
The expected values of the three arms' rewards are shown in Figure \ref{PotentialOutcomes}. 

In the warm-start data, $x_{t,0}$ and $x_{t,1}$ are generated from a truncated normal distribution $\mathcal{N}(0, 1)$ on the interval $(-1.15, -0.85)$, while in subsequent data $x_{t,0}$ and $x_{t,1}$ are drawn from $\mathcal{N}(0, 1)$ without the truncation.
Each one of the 50 warm-start contexts is assigned to one of the three arms at random with equal probability. 
Note that the warm-start contexts belong to a region of the context space where the reward surfaces do not change much with the context. 
Therefore, when training the reward model for the first time, the estimated reward of arm $a = 2$ (blue) is the highest, the one of arm $a = 1$ (yellow) is the second highest and the one of arm $a = 0$ (red) is the lowest across the context space. 

We run our experiment with a learning horizon $T = 10000$. The regularization parameter $\lambda$, which is present in all algorithms, is chosen via cross-validation every time the model is updated. The constant $\alpha$, which is present in all algorithms, is optimized among values $0.25, 0.5, 1$ in the Thompson sampling bandits (the value $\alpha=1$ corresponds to standard Thompson sampling, \cite{chapelle-tsucb} suggest that smaller values may lower regret) and among values $1, 2, 4$ in the UCB bandits \cite{chapelle-tsucb}. The propensity threshold $\gamma$ for BLTS and BLUCB is optimized among the values $0.01, 0.05, 0.1, 0.2$.

\subsubsection{Well-Specified Outcome Models}\label{well}

In this section, we compare the behavior of LinTS, LinUCB, BLTS and BLUCB when the outcome model of the contextual bandits is well-specified, i.e., it includes both linear and quadratic terms.
Note that this is still in the domain of linear contextual bandits, if we treat the quadratic terms as part of the context.

\begin{figure}[t]	
	\centering
	\subfloat[Well-specified LinTS]{\includegraphics[width=0.66\textwidth]{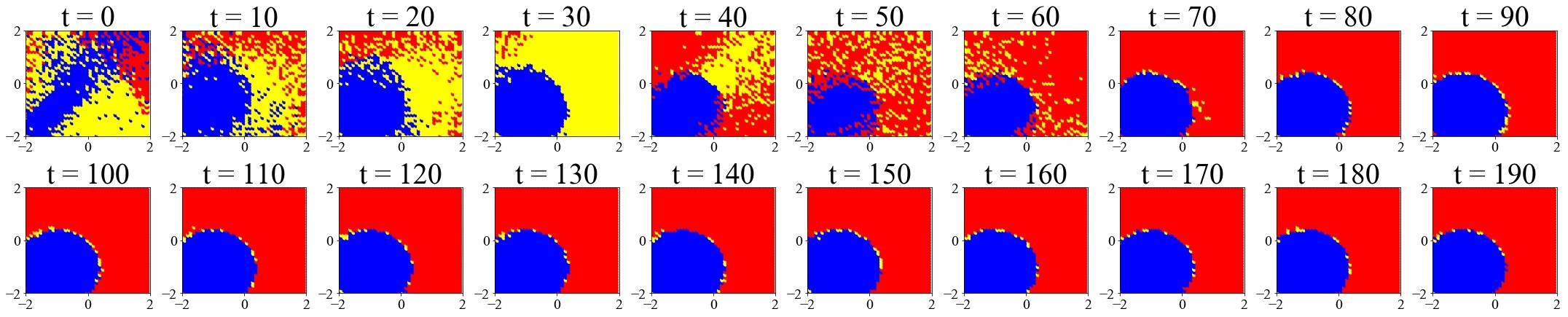}
		\label{RidgeTS-SecondOrder-PWFalse}} \\
	
	\subfloat[Well-specified LinUCB]{\includegraphics[width=0.66\textwidth]{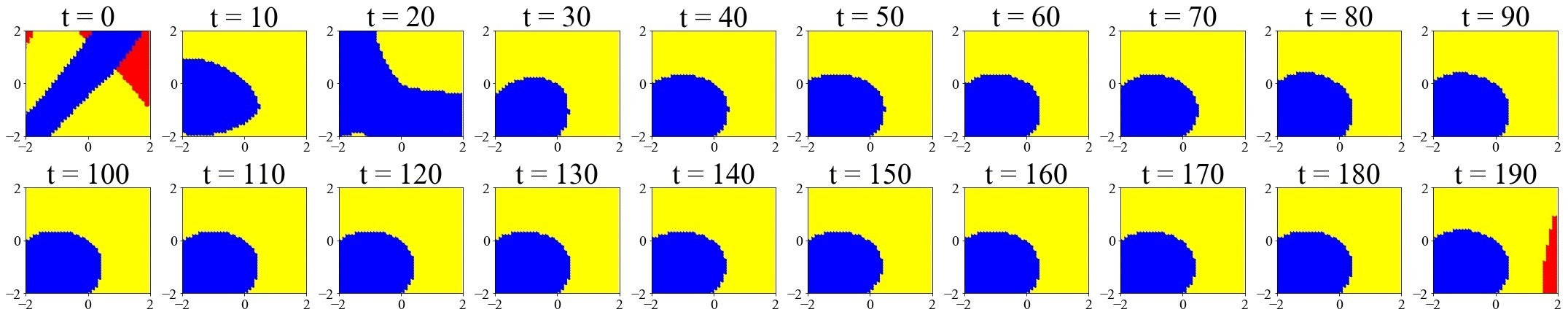}
		\label{RidgeUCB-SecondOrder-PWFalse}} \\
	\subfloat[Well-specified BLTS]{\includegraphics[width=0.66\textwidth]{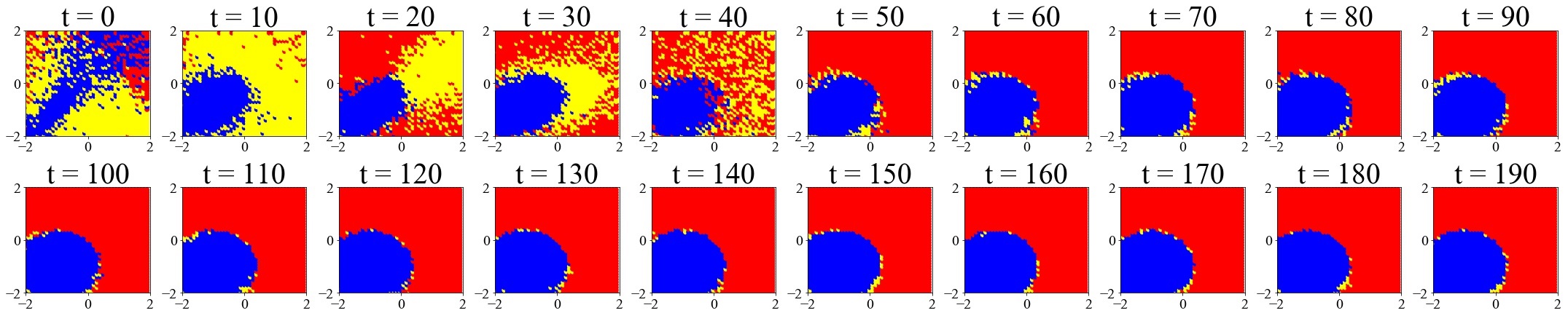}
		\label{RidgeTS-SecondOrder-PWTrue}}
	
	\subfloat[Well-specified BLUCB]{\includegraphics[width=0.66\textwidth]{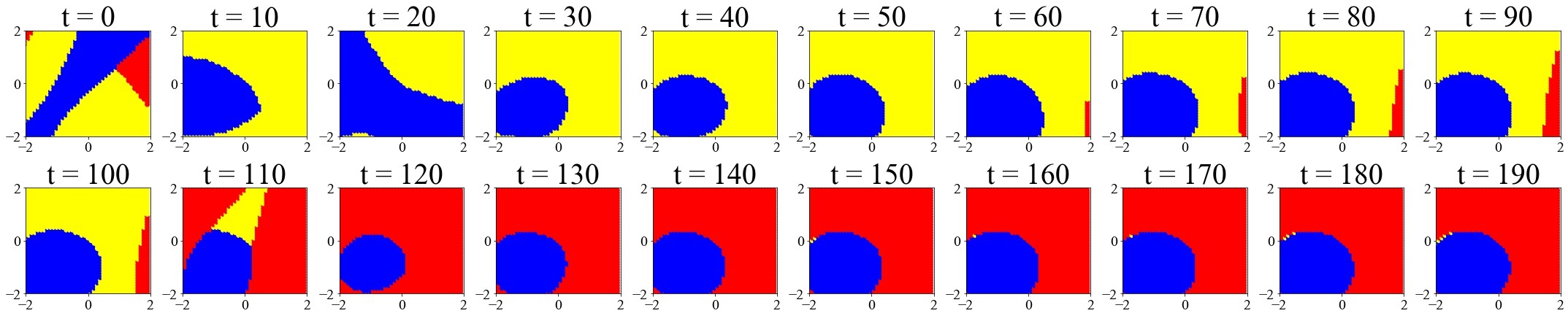}
		\label{RidgeUCB-SecondOrder-PWTrue}} \\
	\caption{Evolution of the arm assignment in the context space for well-specified LinTS, LinUCB, BLTS, BLUCB.}
\end{figure}

First, we compare LinTS and LinUCB. 
Figure \ref{RidgeTS-SecondOrder-PWFalse} shows that 
the uncertainty and the stochastic nature of LinTS leads to a ``dispersed'' assignment of arms $a=1$ and $a=2$ and to the crucial assignment of a few contexts to arm $a = 0$. 
This allows LinTS to start decreasing the bias in the estimation of all three arms in the subsequent time periods.
Within the first few learning observations, LinTS estimates the outcome models of all three arms correctly and finds the optimal assignment. 
On the other hand, Figure \ref{RidgeUCB-SecondOrder-PWFalse}, shows that the deterministic nature of LinUCB assigns entire regions of the context space to the same arm. 
As a result not enough contexts are assigned to $a=0$ and LinUCB delays the correction of bias in the estimation of this arm.  
Another way to understand the problem is that the outcome model in the LinUCB bandit has biased coefficients combined with estimated uncertainty that is too low to incentivize the exploration of arm $a = 0$ initially. 
LinUCB finds the correct assignment after 240 observations.

\begin{figure}[!htb]
	\centering
	\includegraphics[width=0.8\textwidth]{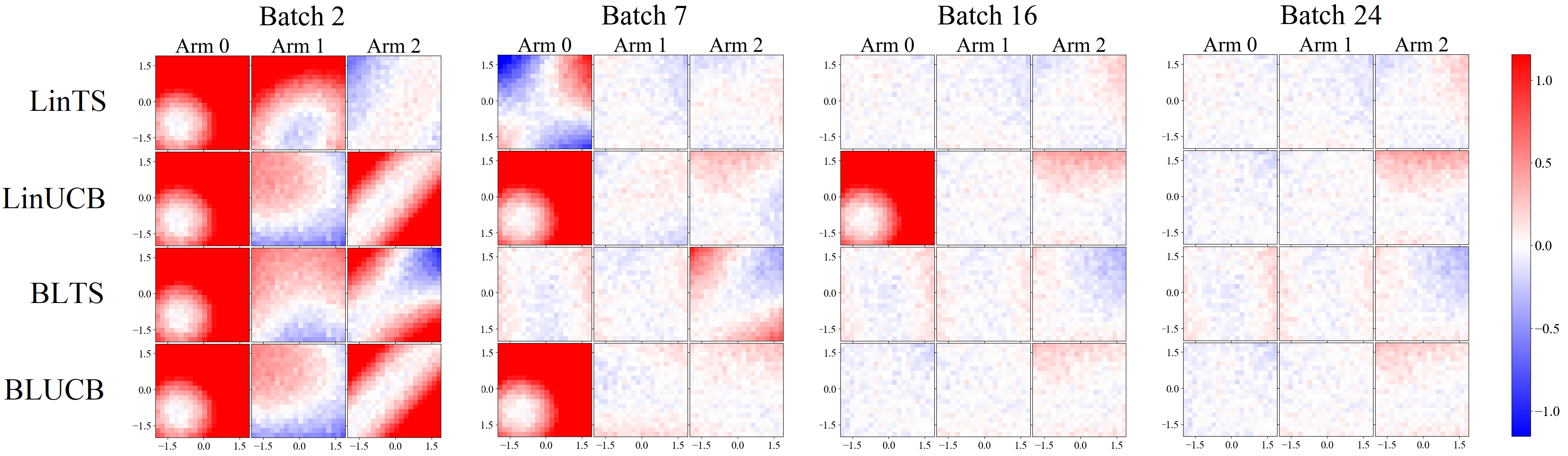}
	\caption{Evolution of the potential outcomes estimation bias in the $(x_0, x_1)$ context space for well-specified LinTS, LinUCB, BLTS and BLUCB. Blue indicates that the actual estimate is lower than the predicted, whereas red indicates that the actual estimate is higher than the predicted.}
	\label{RidgeTSUCB-SecondOrder-Bias}
\end{figure}

Second, we study the performance of  BLTS and BLUCB.
In Figure \ref{RidgeUCB-SecondOrder-PWTrue}, we observe that balancing has a significant impact on the performance of UCB, since BLUCB finds the optimal assignment after 110 observations, much faster than LinUCB.
This is because the few observations of arm $a = 0$ outside of the context region of the warm-start batch are weighted more heavily by BLUCB. 
As a result, BLUCB, despite its deterministic nature which complicates estimation, is able to reduce its bias more quickly via balancing 
Figure \ref{RidgeTS-SecondOrder-PWTrue} shows that BLTS is also able to find the optimal assignment a few observations earlier than LinTS.
Figure \ref{RidgeTSUCB-SecondOrder-Bias} shows the evolution of the estimation bias for all three arms for the well-specified LinTS, LinUCB, BLTS and BLUCB.


The first column of Table \ref{RidgeTSUCB-Percentages} shows the percentage of simulations in which LinTS, LinUCB, BLTS and BLUCB find the optimal assignment within $T = 10000$ contexts for the well-specified case.
BLTS outperforms all other algorithms by a large margin.

\subsubsection{Mis-Specified Outcome Models}

We now study the behavior of LinTS, LinUCB, BLTS and BLUCB when the outcome models include only linear terms of the context and therefore are mis-specified.
In real-world domains, the true data generative process is complex and very difficult to capture by the simpler outcome models assumed by the learning algorithms. 
Hence, model mismatch is very likely. 

\begin{figure}[t]	
	\centering
	\subfloat[Mis-specified LinTS]{\includegraphics[width=0.66\textwidth]{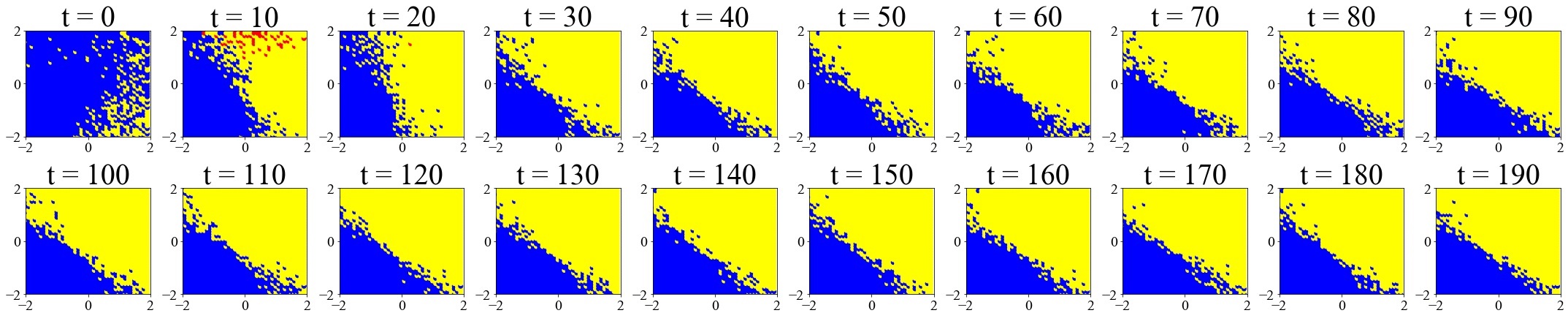}
		\label{RidgeTS-FirstOrder-PWFalse}} \\
	\subfloat[Mis-specified LinUCB]{\includegraphics[width=0.66\textwidth]{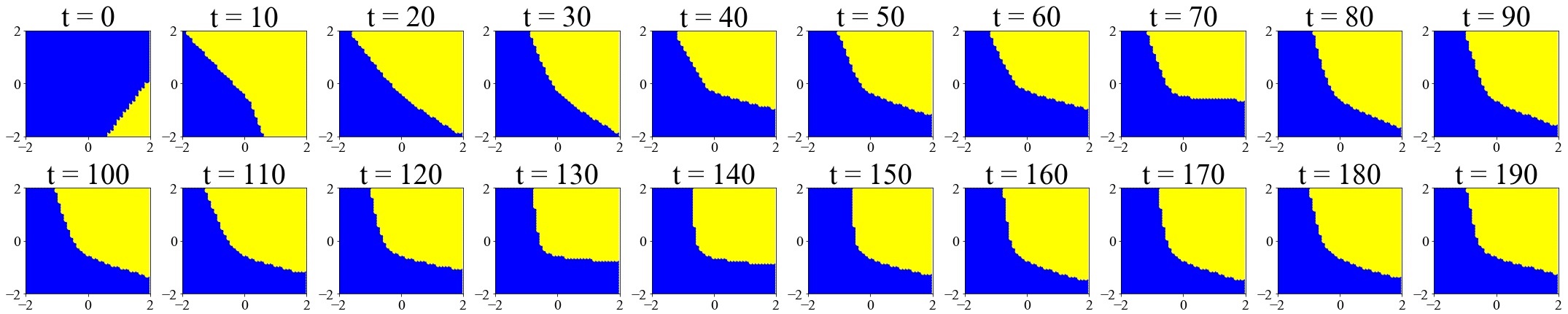}
		\label{RidgeUCB-FirstOrder-PWFalse}} \\
	\subfloat[Mis-specified BLTS]{\includegraphics[width=0.66\textwidth]{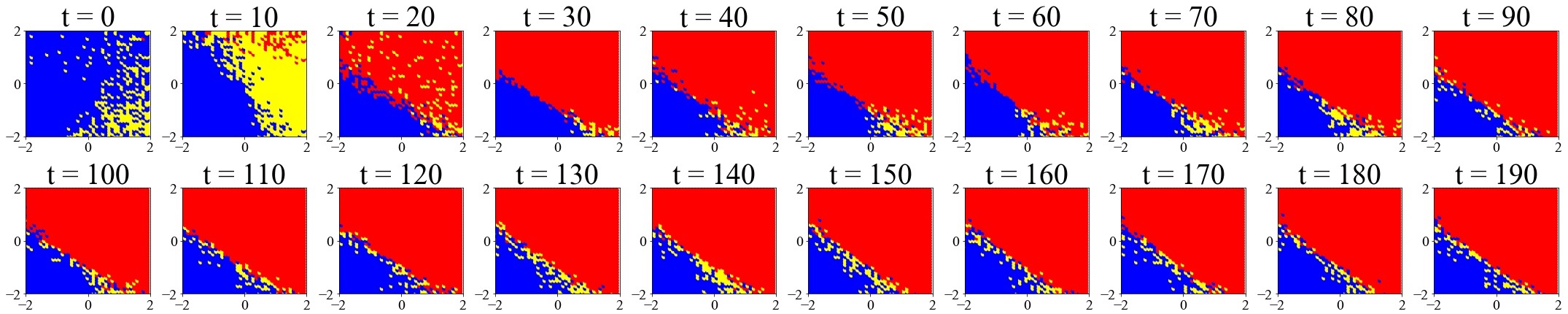}
		\label{RidgeTS-FirstOrder-PWTrue}} \\
	\subfloat[Mis-specified BLUCB]{\includegraphics[width=0.66\textwidth]{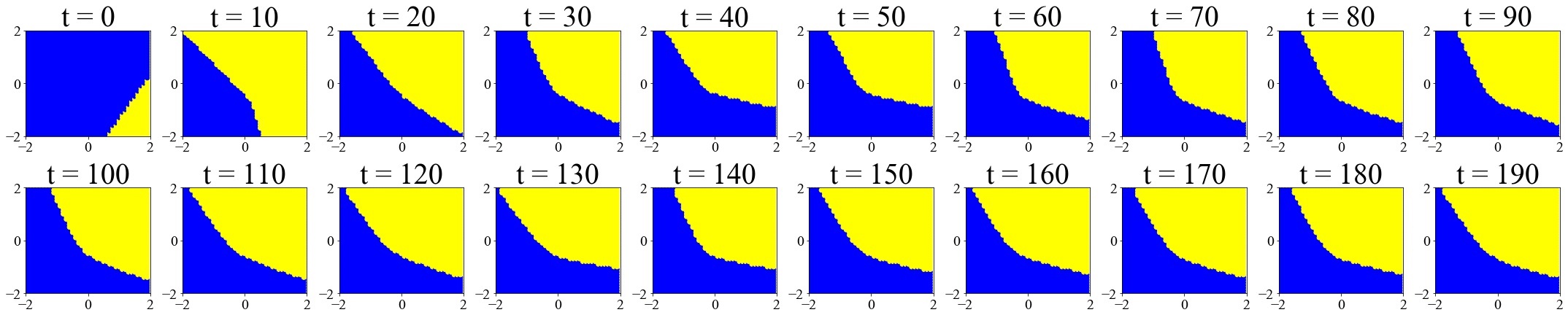}
		\label{RidgeUCB-FirstOrder-PWTrue}}
	\caption{Evolution of the arm assignment in the context space for mis-specified LinTS, LinUCB, BLTS, BLUCB.}
\end{figure}

\begin{figure}[!htb]
	\centering
	\includegraphics[width=0.8\textwidth]{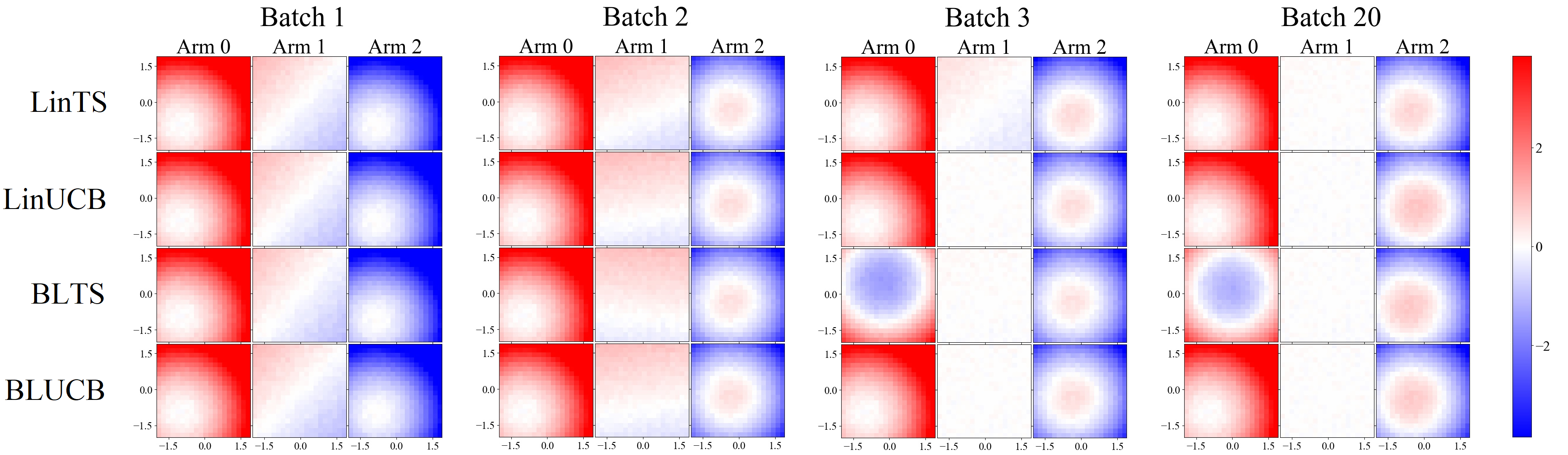}
	\caption{Evolution of the potential outcomes estimation bias in the $(x_0, x_1)$ context space for mis-specified LinTS, LinUCB, BLTS and BLUCB. Blue indicates that the actual estimate is lower than the predicted, whereas red indicates that the actual estimate is higher than the predicted.}
	\label{RidgeTSUCB-FirstOrder-Bias}
\end{figure}

We first compare LinTS and LinUCB.
In Figures \ref{RidgeTS-FirstOrder-PWFalse}, \ref{RidgeUCB-FirstOrder-PWFalse}, we see that during the first time periods, both bandits assign most contexts to arm $a = 2$ and a few contexts to arm $a = 1$. 
LinTS finds faster than LinUCB the linearly approximated area in which arm $a = 2$ is suboptimal. 
However, both LinTS and LinUCB have trouble identifying that the optimal arm is $a = 0$. 
Due to the low estimate of $a = 0$ from the mis-representative warm-start observations, LinUCB does not assign contexts to arm $a = 0$ for a long time and therefore, delays to estimate the model of $a = 0$ correctly. 
LinTS does assign a few contexts to arm $a = 0$, but they are not enough to quickly correct the estimation bias of arm $a = 0$ either. 
On the other hand, BLTS is able to harness the advantages of the stochastic assignment rule of Thompson sampling. 
The few contexts assigned to arm $a = 0$ are weighted more heavily by BLTS. 
Therefore, as shown in Figure \ref{RidgeTS-FirstOrder-PWTrue}, BLTS corrects the estimation error of arm $a = 0$ and finds the (constrained) optimal assignment already after 20 observations. 
On the other hand, BLUCB does not handle better than LinUCB the estimation problem created by the deterministic nature of the assignment in the mis-specified case, as shown in Figure \ref{RidgeUCB-FirstOrder-PWTrue}.
Figure \ref{RidgeTSUCB-FirstOrder-Bias} shows the evolution of the estimation bias for all three arms for the mis-specified LinTS, LinUCB, BLTS and BLUCB.

%
The second column of table \ref{RidgeTSUCB-Percentages} shows the percentage of simulations in which LinTS, LinUCB, BLTS and BLUCB find the optimal assignment within $T = 10000$ contexts for the mis-specified case.
Again, BLTS has a strong advantage.


\vspace*{12pt}
This simple synthetic example allowed us to explain transparently where the benefits of balancing in linear bandits stem from. Balancing helps escape biases in the training data and can be more robust in the case of model mis-specification. 
While, as we proved, balanced linear contextual bandits share the same strong theoretical guarantees, this indicates towards their better performance in practice compared to other contextual bandits with linear realizability assumption.
We investigate this further in the next section with an extensive evaluation on real cost-sensitive classification datasets.

\begin{table}[h]
	\centering
	\begin{tabular}{l||c||c}
		& Well-Specified & Mis-Specified \\
		\hline 
		LinTS & 84\% & 39\% \\ 
		\hline
		LinUCB & 51\% & 29\%  \\ 
		\hline
		{BLTS} & {92\%} & {58\%}  \\ 
		\hline
		BLUCB & 79\% & 30\% \\
		\hline
	\end{tabular}
	\caption{Percentage of simulations in which LinTS, LinUCB, BLTS and BLUCB find the optimal assignment within learning horizon of $10000$ contexts}
	\label{RidgeTSUCB-Percentages}
\end{table}

\subsection{Multiclass Classification with Bandit Feedback}\label{subsec:multi_class}

Adapting a classification task to a bandit problem is a common method for comparing contextual bandit algorithms \cite{dudik-offline-1}, \cite{agarwal-ilovetoconbandits}, \cite{bietti2018contextual}. In a classification task, we assume data are drawn IID from a fixed distribution: $(x, c) \sim D$, where $x \in \X$ is the context and $c \in {1, 2, \dots, K}$ is the class. The goal is to find a classifier $\pi: \X \rightarrow \{1, 2, \dots, K\}$ that minimizes the classification error $\E_{(x, c) \sim D}\mathbf{1}\left\{\pi(x) \neq c\right\}$. The classifier can be seen as an arm-selection policy and the classification error is the policy's expected regret. Further, if only the loss associated with the policy's chosen arm is revealed, this becomes a contextual bandit setting. So, at time $t$, context $x_t$ is sampled from the dataset, the contextual bandit selects arm $a_t \in \{1, 2, \dots, K\}$ and observes reward $r_t(a_t) = \mathbf{1}\left\{a_t = c_t\right\}$, where $c_t$ is the unknown, true class of $x_t$. 
The performance of a contextual bandit algorithm on a dataset with $n$ observations is measured with respect to the normalized cumulative regret, $\frac{1}{n} \sum_{t =1}^n{\left(1 - r_t(a_t)\right)}$.

We use 300 multiclass datasets from the Open Media Library (OpenML). The datasets vary in number of observations, number of classes and number of features. Table \ref{stats} summarizes the characteristics of these benchmark datasets. Each dataset is randomly shuffled. 

\begin{table}[H]
	\centering
	\begin{tabular}{|c|c|}
		\hline
		Observations & Datasets \\
		\hline 
		$\leq 100$ & 58 \\
		\hline
		$> 100$ and $\leq 1000$  & 152  \\ 
		\hline
		$> 1000$ and $\leq 10000$ & 57  \\ 
		\hline
		$> 10000$ & 33 \\
		\hline
	\end{tabular}
	\hspace{10pt}
	\begin{tabular}{|c|c|}
		\hline
		Classes & Count \\
		\hline 	
		$2$ & 243 \\
		\hline
		$> 2 \text{ and } 10$  & 48  \\ 
		\hline
		$ > 10 $ & 9  \\
		\hline
	\end{tabular}
	\hspace{10pt}
	\begin{tabular}{|c|c|}
		\hline
		Features & Count \\
		\hline 	
		$\leq 10$ & 154 \\
		\hline
		$> 10 \text{ and } \leq 100$  & 106  \\ 
		\hline
		$> 100$  & 40  \\
		\hline
	\end{tabular}
	\caption{Characteristics of the 300 datasets used for the experiments of multiclass classification with bandit feedback.}
	\label{stats}
\end{table}

We evaluate LinTS, BLTS, LinUCB and BLUCB on these 300 benchmark datasets. We run each contextual bandit on every dataset for different choices of input parameters. The regularization parameter $\lambda$, which is present in all algorithms, is chosen via cross-validation every time the model is updated. The constant $\alpha$, which is present in all algorithms, is optimized among values $0.25, 0.5, 1$ in the Thompson sampling bandits \cite{chapelle-tsucb} and among values $1, 2, 4$ in the UCB bandits \cite{chapelle-tsucb}. The propensity threshold $\gamma$ for BLTS and BLUCB is optimized among the values $0.01, 0.05, 0.1, 0.2$.
Apart from baselines that belong in the family of contextual bandits with linear realizability assumption and have strong theoretical guarantees, we also evaluate the policy-based ILOVETOCONBANDITS (ILTCB) from \cite{agarwal-ilovetoconbandits} that does not estimate a model, but instead it assumes access to an oracle for solving fully supervised cost-sensitive classification problems and achieves the statistically optimal regret.

\begin{figure}[H]
	\centering
	\includegraphics[width=0.95\linewidth]{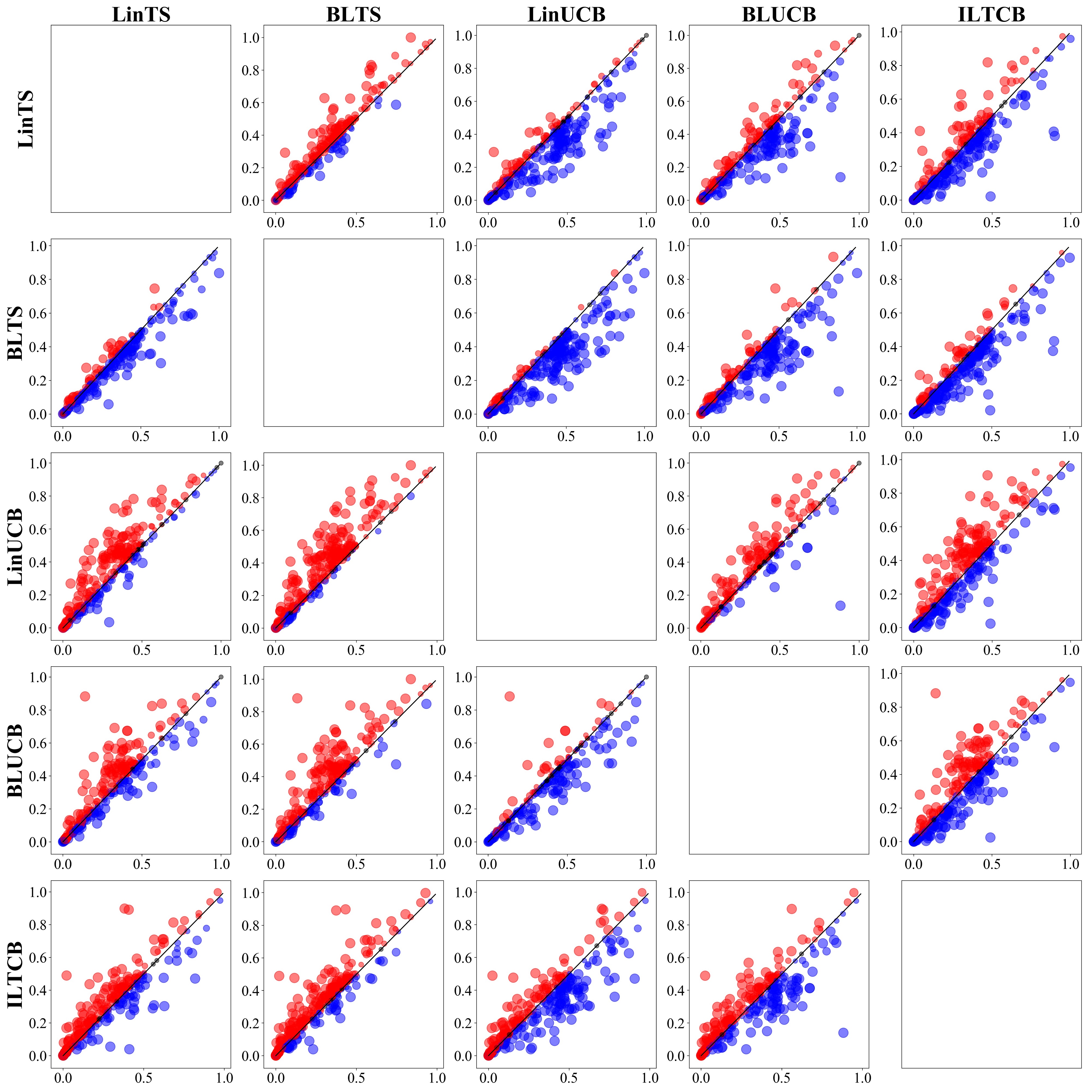}
	\caption{Pairwise comparison of LinTS, BLTS, LinUCB, BLUCB on 300 classification datasets. BLUCB outperforms LinUCB. BLTS outperforms LinTS, LinUCB, BLUCB.}
	\label{fig:all_vs_all}
\end{figure}

Figure \ref{fig:all_vs_all} shows the pairwise comparison of LinTS, BLTS, LinUCB, BLUCB and ILTCB on the 300 classification datasets. Each point corresponds to a dataset. The $x$ coordinate is the normalized cumulative regret of the column bandit and the $y$ coordinate is the normalized cumulative regret of the row bandit. The point is blue when the row bandit has smaller normalized cumulative regret and wins over the column bandit. The point is red when the row bandit loses from the column bandit. The point's size grows with the significance of the win or loss.

The first important observation is that the improved model estimation achieved via balancing leads to better practical performance across a large number of contextual bandit instances. Specifically, BLTS outperforms LinTS and BLUCB outperforms LinUCB.
The second important observation is that deterministic assignment rule bandits are at a disadvantage compared to randomized assignment rule bandits. The improvement in estimation via balancing is not enough to outweigh the fact that estimation is more difficult when the assignment is deterministic and BLUCB is outperformed by LinTS.
Overall, BLTS which has both balancing and a randomized assignment rule, outperforms all other linear contextual bandits with strong theoretical guarantees.
BLTS also outperforms the model-agnostic ILTCB algorithm.

We refer the reader to Appendix B of the supplemental material for details on the datasets.

\subsection{Theoretical Guarantee}\label{subsec:guarantee}

Here we establish theoretical guarantees of BLTS and BLUCB that are comparable to LinTS and LinUCB. We start with a few technical assumptions that are standard in the contextual bandits literature.

\begin{assumption}\label{assump:realizability}
	\textbf{Linear Realizability:}
	There exist parameters $\{\theta_a\}_{a\in \A}$ such that given any context $x$, $\mathbb{E}[r_t(a) | x] = x^\top \theta_a, \forall a \in \A, \forall t \ge 0$.
\end{assumption}

We use the standard (frequentist) regret criterion to measure performance as defined next:
\begin{definition}
	The instantaneous regret at iteration $t$ is $x^\top_t \theta_{a_t^*} -  x^\top_t \theta_{a_t}$, 
	where $a^*_t$ is the optimal arm at iteration $t$ and $a_t$ is the arm taken at iteration $t$.
	The cumulative regret $R(T)$ with horizon $T$ is the defined as $R(T) = \sum_{t=1}^\top \left(x^\top_t \theta_{a_t^*} -  x^\top_t \theta_{a_t}\right)$.
\end{definition}

\begin{definition}
	We denote the canonical filtration of the underlying contextual bandits problem by $\{\mathcal{F}_t\}_{t=1}^{\infty}$, where $\mathcal{F}_t = \sigma(\{x_s\}_{s=1}^t, \{a_s\}_{s=1}^t, \{r_s(a_s)\}_{s=1}^t, x_{t+1})$: the sigma algebra\footnote{All the random variables $x_t, a_t, r_t$ are defined on some common underlying probability space, which we do not write out explicitly here.} generated by all the random variables up to and including iteration $t$, plus $x_{t+1}$. In other words, $\mathcal{F}_t$ contains all the information that is available before making the decision for iteration $t+1$. 
\end{definition}

Next, we make the standard assumptions on the regularity of the distributions: 
\begin{assumption}\label{assump:reg}
	For each $a \in \A$ and every $t \ge 1$:
	\begin{enumerate}
		\item 
		\textbf{Sub-Gaussian Noise:}
		$r_t(a) - x_t^\top\theta_a$ is conditionally sub-Gaussian:
		there exists some $L_a > 0$, such that
		$\mathbb{E} [e^{s(r_t(a) - x_t^\top\theta_a)} \mid \mathcal{F}_{t-1}] \le \exp(\frac{s^2 L_a^2}{2}), \forall s, \forall x_t$.
		\item
		\textbf{Bounded Contexts and Parameters:}
		The contexts $x_t$ and parameters $\theta_a$ are assumed to be bounded.
		Consequently, without loss of generality, we can rescale them such that
		$\|x_t\|_2 \le 1, \|\theta_a\|_2 \le 1, \forall a, t$.
	\end{enumerate}
	
\end{assumption}

\begin{remark}
	Note that we make no assumption of the underlying $\{x_t\}_{t=1}^{\infty}$ process:
	the contexts $\{x_t\}_{t=1}^{\infty}$ need not to be fixed beforehand or come from some stationary process. Further, they can even be adapted to
	$\sigma(\{x_s\}_{s=1}^t, \{a_s\}_{s=1}^t, \{r_s(a_s)\}_{s=1}^t)$, in which case they are called adversarial contexts in the literature as the contexts can be chosen by an adversary who chooses a context after observing the arms played and the corresponding rewards.
	If $\{x_t\}_{t=1}^{\infty}$ is an IID process, then the problem is known as stochastic contextual bandits. From this viewpoint, adversarial contextual bandits are more general, but the regret bounds tend to be worse. Both are studied in the literature.
\end{remark}

\begin{theorem}\label{thm:main}
	Under Assumption~\ref{assump:realizability} and Assumption~\ref{assump:reg}:
	\begin{enumerate}
		\item 
		If BLTS is run with $\alpha = \sqrt{\frac{\log\frac{1}{\delta}}{\epsilon}}$ in Algorithm \ref{alg:BLTS},
		then with probability at least $1 - \delta$, 
		$R(T) = \tilde{O}\left(d\sqrt{\frac{KT^{1+\epsilon}}{\epsilon}}\right)$.
		\item If BLUCB is run with $\alpha = \sqrt{\log \frac{TK}{\delta}}$ in Algorithm \ref{alg:BLUCB},
		then with probability at least $1 - \delta$, 
		$R(T) = \tilde{O}\left(\sqrt{TdK}\right)$.
	\end{enumerate}
\end{theorem}

\noindent We refer the reader to Appendix A of the supplemental material for the regret bound proofs.

\begin{remark}
	The above bound essentially matches the existing state-of-the art regret bounds for linear Thompson sampling with direct model estimation (e.g.~\cite{agrawal-lints}). Note that in~\cite{agrawal-lints},
	an infinite number of arms is also allowed, but all arms share the same parameter $\theta$. The final regret bound is $\tilde{O}\left(d^2 \frac{\sqrt{T^{1+\epsilon}}}{\epsilon}\right)$. Note that even though no explicit dependence on $K$ is present in the regret bound (and hence our regret bound appears as a factor of $\sqrt{K}$ worse), this is to be expected, as we have $K$ parameters to estimate, one for each arm. Note that here we do not assume any structure on the $K$ arms; they are just $K$ stand-alone parameters, each of which needs to be independently estimated. 
	Similarly, for BLUCB, our regret bound is $\tilde{O}\left(\sqrt{TdK}\right)$,
	which is a factor of $\sqrt{K}$ worse than that of \cite{chu2011contextual},
	which establishes a $\tilde{O}\left(\sqrt{Td}\right)$ regret bound.
	Again, this is because a single true $\theta^*$ is assumed in \cite{chu2011contextual}, rather than $K$ arm-dependent parameters.
	
	Of course, we also point out that our regret bounds are not tight, nor do they achieve state-of-the-art regret bounds in contextual bandits algorithms in general.
	The lower bound $\Omega({\sqrt{dT}})$ is established in~\cite{chu2011contextual} for linear contextual bandits (again in the context of a single parameter $\theta$ for all $K$ arms).
	In general, UCB based algorithms (\cite{auer-linrel,chu2011contextual,bubeck2012regret,abbasi2011improved}) tend to have better
	(and sometimes near-optimal) theoretical regret bounds.
	In particular, the state-of-the-art bound of $O(\sqrt{dT\log K})$ for linear contextual bandits is given 
	in~\cite{bubeck2012regret} (optimal up to a $O(\log K)$ factor).
	However, as mentioned in the introduction, Thompson sampling based algorithms tend to perform much better in practice (even though their regret bounds tend not to match UCB based algorithms, as is also the case here). Hence, our objective here is not to provide
	state-of-the-art regret guarantees. Rather, we are motivated to design algorithms that have better empirical performance (compared to both the existing UCB style algorithms and Thompson sampling style algorithms), which also enjoy the baseline theoretical guarantee.
	
	Finally, we give some quick intuition for the proof. For  BLTS, we first show that estimated means concentrate around true mean (i.e. $x_t^\top \hat{\theta}_a$ concentrates around $x_t^\top \theta_a$). Then, we establish that sampled means concentrate around the estimated means (i.e. $x_t^\top\tilde{\theta}_a$  concentrates around $x_t^\top\hat{\theta}_a$). These two steps together indicate that the sampled mean is close to the true mean. A further consequence of that is we can then bound the instantaneous regret (regret at each time step $t$) in terms of the sum of two standard deviations: one corresponds to the optimal arm at time $t$, the other corresponds to the actual selected arm at $t$. The rest of the proof then follows by giving tight characterizations of these two standard deviations. For BLUCB, the proof again utilizes the first concentration mentioned above: the estimated means concentrate around true mean (note that there is no sampled means in BLUCB). The rest of the proof adopts a similar structure as in~\cite{chu2011contextual}.
\end{remark}

\section{Additional Estimation Considerations}\label{sec:other_considerations}
In this section, we investigate three additional estimation considerations and present further evidence from simulations that demonstrate our design outlined in Section~\ref{sec:designs} can be more effective in a variety of different contexts. 

\subsection{The Effect of Simpler Outcome Models: LASSO v.s. Ridge}\label{subsec:ridgevslasso}
In contextual bandits, the assignment to an arm is a function of the contexts and not all contexts have the same assignment probabilities. 
This creates an environment with confounding, often in combination with small samples, in which the choice of the estimation method plays a very significant role.
In this section, we study the significance maintaining a simple outcome model in contextual bandits. 
The potential outcome models corresponding to each arm are estimated in every batch and are used by the bandit to determine the assignment of contexts to arms, e.g. via the construction of upper confidence bounds. 
Therefore, maintaining a simple outcome model results in a simple assignment model.  

We compare a contextual bandit that maintains a simpler outcome model, such as LASSO, with a contextual bandit that maintains a more complex outcome model, such as ridge. 
As mentioned in \ref{sec:designs}, exact computation of the reward mean and variance in the case of LASSO  is intractable and we must perform approximation \cite{russo2018tutorial}.
Bootstraping provides a viable way to approximate the quantities. More specifically, we can obtain a sampling distribution on $\mu_a(x)$ by training many regularized regression models on bootstrap samples drawn from $(\textbf{X}_a, \textbf{r}_a)$. With this sampling distribution, we can then easily to compute the mean estimate $\hat{\mu}_\w(\x)$ and the variance estimate $\hat{\sigma}_\w^2(\x)$.
This section compares bootstrap LASSO Thompson sampling and bootstrap ridge Thompson sampling.
In Appendix C, we provide a Bayesian way of using LASSO estimation in a Thompson sampling or UCB contextual bandit motivated by \cite{park2008bayesian}. The theoretical analysis of Bayesian LASSO contextual bandit may be proven more straightforward than the theoretical analysis of bootstrap LASSO contextual bandit, though we leave this analysis for future work.

Consider a simulation setting where the contexts $x_t$ are $d$-dimensional and $x_t \sim \mathcal{N}(0_d, I_{d})$. 
There are three arm $\A = \{0, 1, 2\}$ and the potential outcomes are generated as $r_t(0) = x_{t,0} + f(x_t) + \epsilon_t$, $r_t(1) = 1 - x_{t,0} + f(x_t) + \epsilon_t$, and $r_t(2) = x_{t,1} + f(x_t) +\epsilon_t$, where $\epsilon_t \sim \mathcal{N}(0, \sigma^2)$ and $f(x_t)$ is a function that shifts the potential outcomes of all arms. 
Therefore, in this design, only contextual variables $x_{t,0}$ and $x_{t,1}$ are relevant to the  assignment model. 
Among the remaining $d-2$ contextual variables, there are $q$ nuisance contextual variables that appear in the outcome model, but should not play a role in the assignment model, as their effect to the potential outcomes is the same for all arms. 
These nuisance contextual variables shift the potential outcomes  by $f(x_t) = 2 \sigma \sum_{j = 2}^{q + 1} x_{t,j}$. 
The remaining $d-q-2$ contextual variables are noise contextual variables and do not play a role in either the assignment or the outcome model.

\begin{figure}[h]
\subfloat[MSE for $\w = 0$]{\includegraphics[width=0.32\textwidth]{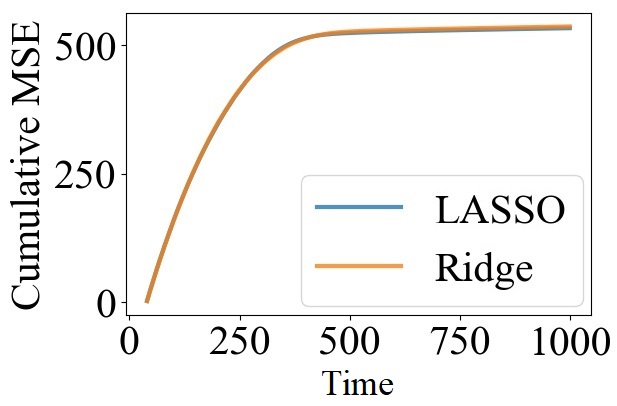}
\label{LR-MSE0-Only}}
\subfloat[MSE for $\w = 1$]{\includegraphics[width=0.32\textwidth]{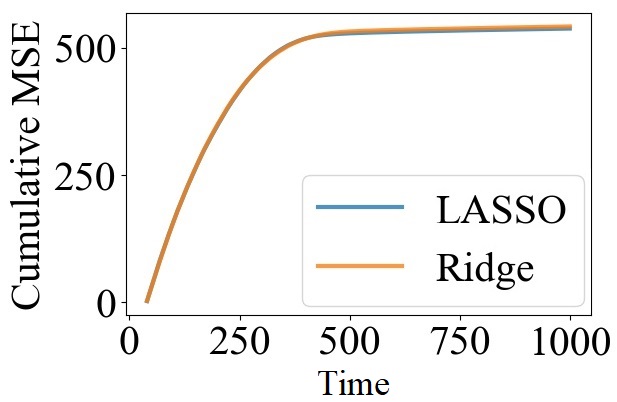}
\label{LR-MSE1-Only}}
\subfloat[MSE for $\w = 2$]{\includegraphics[width=0.32\textwidth]{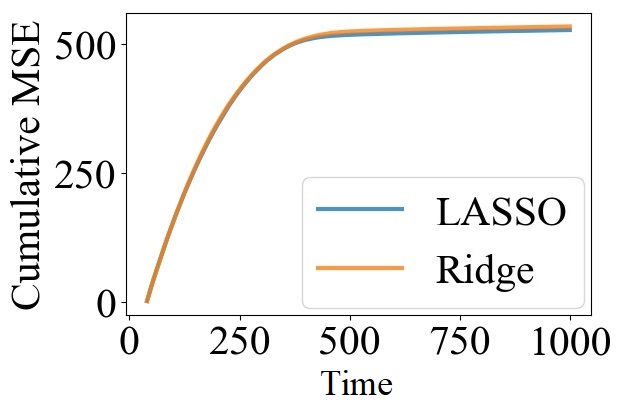}
\label{LR-MSE2-Only}}	
\caption{Cumulative MSE averaged over hundreds of simulations for LASSO and ridge of all arms' outcome model estimation on 1000 randomized observations.}
\label{LR-train}
\end{figure}

We choose $d = 102$ and $q = 51$. 
So, among the 100 non-relevant contextual variables, 51 are nuisance contextual variables and 49 are noise contextual variables. 
First, we compare LASSO and ridge in terms of fit on 1000 observations, when the assignment of contexts to arms is purely randomized. 
The number of nuisance contextual variables, $q$, is chosen so that LASSO and ridge perform equivalently in terms of fit on purely randomized data, in order to evaluate the choice of the estimation method in the adaptive setting all else being equal. 
Indeed, as demonstrated in Figure \ref{LR-train}, the LASSO and ridge models are almost identical in terms of mean squared error (MSE) when trained on batches of randomized data. 

\begin{figure}[h]
\centering
\includegraphics[width=0.48\textwidth]{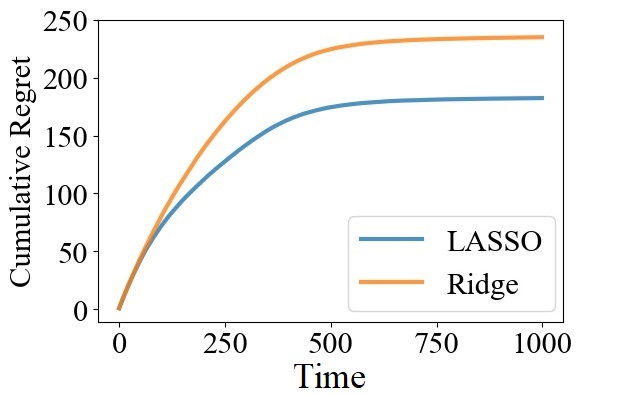}
\caption{Cumulative regret averaged over hundreds of simulations for LASSO Thompson sampling and ridge Thompson sampling on 1000 learning observations.}
\label{LR-bandit}
\end{figure}

Subsequently, we compare the performance of LASSO and ridge in the bandit setting with learning horizon 1000 context. 
Despite the initial equivalence of LASSO and ridge in the training setting, LASSO clearly outperforms ridge in terms of regret in the adaptive learning setting, as shown in Figure \ref{LR-bandit}. 

\begin{figure}[h]
\centering
\subfloat[Ridge Weak  Coefficients]{\includegraphics[width=0.4\textwidth]{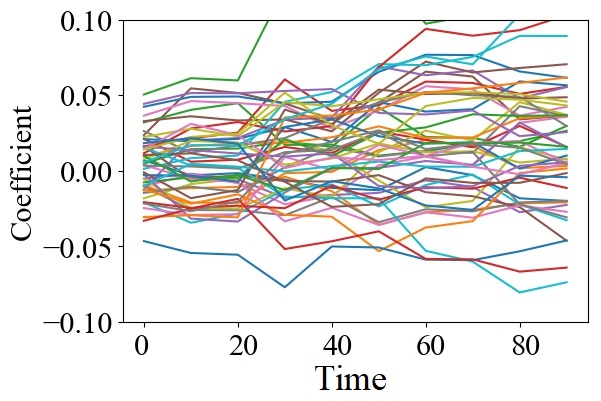}
\label{Ridge-Weak}}
\subfloat[LASSO Weak  Coefficients]{\includegraphics[width=0.4\textwidth]{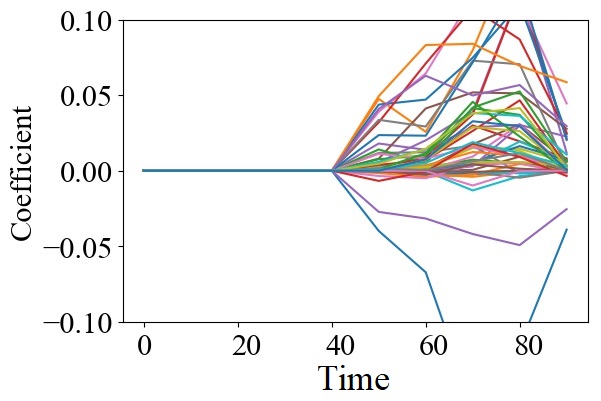}
\label{Lasso-Weak}} \\
\subfloat[Ridge Noise  Coefficients]{\includegraphics[width=0.4\textwidth]{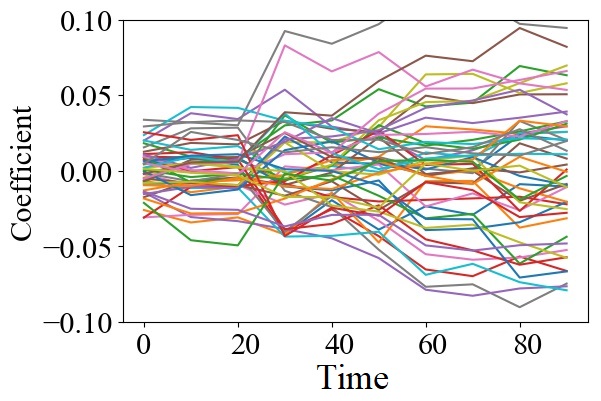}
\label{Ridge-Noise}}
\subfloat[LASSO Noise  Coefficients]{\includegraphics[width=0.4\textwidth]{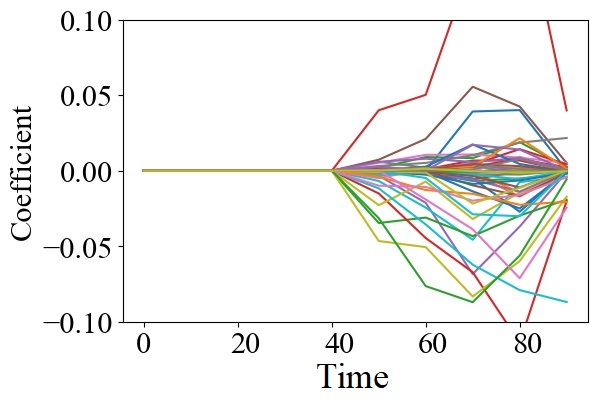}
\label{Lasso-Noise}} 
\caption{Coefficient paths of nuisance and noise contextual variables of the first arm's outcome model in the first 100 learning observations for ridge Thompson sampling and LASSO Thompson sampling.}
\end{figure}

The contributing factor to the bandit performance dissimilarity of these seemingly equivalent models on randomized training data is the presence of confounding. 
In the initial learning observations, a ridge bandit, due to $L_2$ regularization, brings in all of the nuisance and noise contextual variables, as shown in Figures \ref{Ridge-Weak}, \ref{Ridge-Noise}.  
The nuisance contextual variables affect assignment (possibly in non-linear ways) and act as confounders.  
There is insufficient data in the early observations to accurately control for all of them in the outcome model estimation.  
Both the nuisance and the noise contextual variables create more extreme and variable assignment probabilities, increasing the variance of estimation.  
A LASSO bandit, due to the $L_1$ regularization, excludes from the outcome model most of the noise contextual variables and initially, the nuisance contextual variables, as shown in Figures \ref{Lasso-Weak}, \ref{Lasso-Noise}. 
As a result, in the early observations, there are fewer confounders compared to a ridge bandit. 
Therefore, in the subsequent stages of learning, there is less bias as well as less noise in the assignment process for the LASSO bandit than for the ridge bandit.

\subsection{Significance of Non-Parametric Bandits under Non-Linearities}
Real-world applications, such as recommendation systems, may have inherently difficult and complex outcome models. 
In such cases, contextual bandits based on non-parametric model estimation may be more agile. 
In this section, we compare non-parametric contextual bandits (generalized random forest Thompson sampling) with parametric contextual bandits that make linearity assumptions on the potential outcome models (bootstrap LASSO Thompson sampling and bootstrap ridge Thompson sampling).

Consider a simple non-linear simulation design with $10$-dimensional contexts $x_t$ such that $x_{t,j} \sim \mathcal{N}(0, 1), \enskip j = 0, \dots, 9$. 
There are three arms $\A = \{0, 1, 2\}$ and the potential outcomes are generated as $r_t(0) = x_{t,0} + \epsilon_t$, $r_t(1) = - x_{t,0} + \epsilon_t$, and $r_t(2) = \mathbf{1}\{ x_{t,1} < 0\} \left(\min\{x_{t,0}, - x_{t,0}\}- 1\right) + \mathbf{1}\{ x_{t,1} > 0\} \left(\max\{x_{t,0}, - x_{t,0}\} + 1\right) + \epsilon_t$, where $\epsilon_t \sim \mathcal{N}(0, \sigma^2)$ with $\sigma = 0.1$. 
Therefore, only contextual variables $x_{t,0}$ and $x_{t,1}$ are relevant to the arm assignment model. 
In this design, the correct assignment is $\w = 2$ in the first and second quadrants, $\w = 1$ in the third quadrant and $\w = 0$ in the fourth quadrant.
We run the bandits on 5000 observations.
The models of LASSO and ridge include quadratic and second order interarm terms, there are $B = 100$ bootstrap samples, and the regularization parameter is chosen via cross-validation. 
The number of trees in the generalized random forest is $m = 200$ and the tuning parameters are the default, specified in \cite{athey-grf} and in the \verb|grf| R package.

\begin{figure}[h]
\centering
\includegraphics[width=0.5\textwidth]{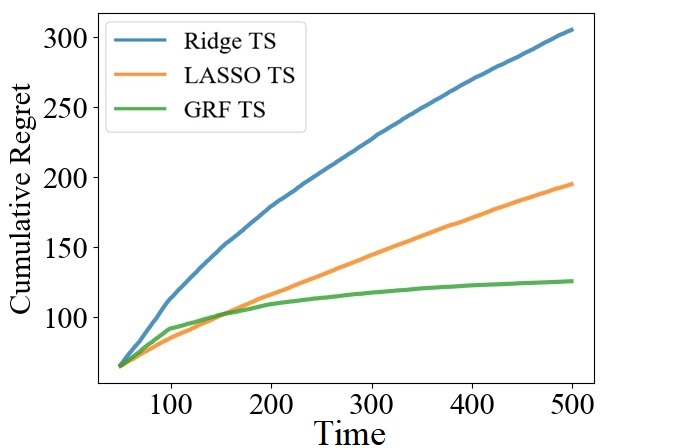}
\caption{Cumulative regret averaged over hundreds of simulations of ridge, LASSO and  generalized random forest Thompson sampling over 5000.}
\label{ParamNonParamRegret}
\end{figure}

\begin{figure}[H]
\subfloat[Ridge Thompson Sampling]{\includegraphics[width=1\textwidth]{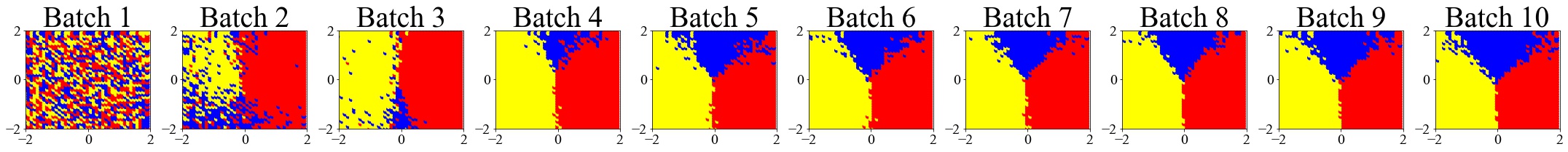}
\label{Tree-RidgeTS}} \\
\subfloat[LASSO Thompson Sampling]{\includegraphics[width=1\textwidth]{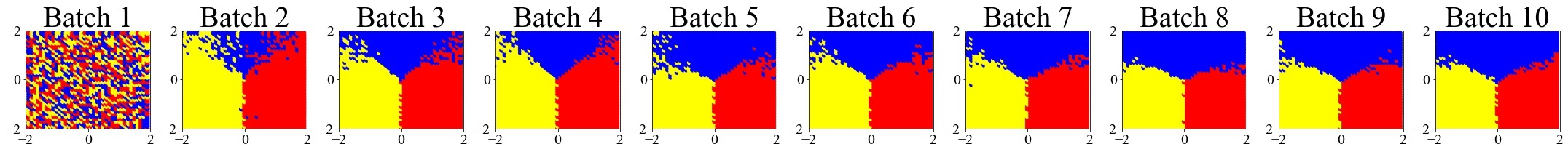}
\label{Tree-LassoTS}} \\
\subfloat[GRF Thompson Sampling]{\includegraphics[width=1\textwidth]{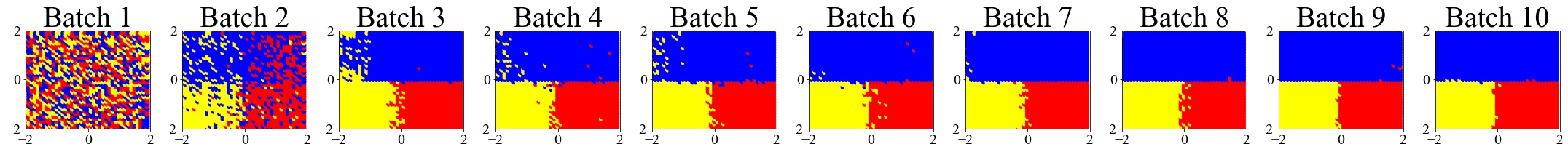}
\label{Tree-GRFTS}}
\caption{Evolution of the arm assignment in the $(x_{\cdot, 0}, x_{\cdot, 1})$ context space for ridge, LASSO and  generalized random forest Thompson sampling. The optimal assignment is $\w = 2$ (blue) in the 1st and 2nd quadrants, $\w = 1$ (yellow) in the 3rd quadrant and $\w = 0$ (red) in the 4th quadrant.}
\label{ParamNonParamAssignment}
\end{figure}

Figure \ref{ParamNonParamRegret} shows that the generalized random forest bandit outperforms the LASSO and the ridge bandits. 
In cases where the outcome functional form is complicated, which is expected in real-world settings, bandits based on non-parametric model estimation may be proven useful and perform better. 
However, one needs to bear in mind that similarly to the ridge bandit, the generalized random forest bandit creates a complex assignment model and is subject to the disadvantages discussed in Section \ref{subsec:ridgevslasso}. 
In this simulation design, the presence of noise contextual variables leads the LASSO bandit to outperform the ridge bandit. 
But the inability of the LASSO bandit and the ability of the generalized random forest bandit to fit the potential outcome model of the third arm, results in a strong performance edge of the latter.  
Figure \ref{ParamNonParamAssignment} shows the assignment evolution in the $(x_{\cdot, 0}, x_{\cdot, 1})$ for the ridge, the LASSO and the generalized random forest bandit.
The generalized random forest bandit has the advantage that the outcome model is non-parametric, and thus is able to account for non-linear functions of the contextual variables, in principle reducing problems that might arise if the assignment model is a non-linear function of contextual variables. 
Additionally, the ``honest'' estimation property featured by the generalized random forest bandit reduces bias.

\subsection{Smoothing the Assignment Policy}

So far, we have considered the assignment rule for each context distinctly; the rule depends on the mean and variance of estimates at each context $x$. 
A literature on optimal policy evaluation in the offline world (e.g. \cite{athey-offline}) derives efficient methods for offline policy estimation when the policy is constrained to be of limited complexity. One example uses trees of limited depth as the relevant policy class. 
The method first constructs the efficient score $\hat{\mu}_{a}(x) + \frac{r - \hat{\mu}_{a}(x)}{\hat{p}_{a}(x)}$ for each observation $(x, a, r)$. 
Subsequently, it estimates a classification tree on the history of contexts $\textbf{X}$ and assignments $\textbf{a}$, weighted by the absolute value of the efficient scores in order to determine the optimal choice of arm in each leaf, where leaves are regions of the context space.  

Here, we propose to follow their method to estimate a policy assignment tree.  
However, we use the output differently: rather than deterministically assigning each context to the estimated optimal policy, instead we use estimates of the mean and variance of each arm within each leaf together with Thompson sampling or UCB to determine assignments.  
One complication that potentially arises in the online setting is that in the early stages of learning the estimation of the ``nuisance parameters'' in the efficient score, $\hat{\mu}_{a}(x)$ and $\hat{p}_{a}(x)$, may be noisy due to the small number of observations.

To understand why using simpler assignment rules through a form of smoothing can be beneficial, it is useful to contrast two cases, one where the probability that an arm is best is estimated very precisely, and the second where we have a noisy estimate of that probability.  
Suppose that sampling according to the true probability balances exploration and exploitation in an ideal way (e.g. that the Thompson sampling heuristic is ideal in a given setting).  
Then, when shifting to the second case where the probabilities are unknown, adding a small amount of smoothing to the assignment rule will have little effect on the exploitation side of the bandit trade-off (given that the estimates were noisy, a little smoothing does not introduce first-order mistakes in assignment).  From a practical perspective, this suggests using simpler assignment rules can improve performance; further, the Thompson sampling heuristic, which incorporates its own form of smoothing by randomizing assignment, may also improve performance over UCB.
However, smoothing improves the exploration side of the trade-off, since it enables lower-variance estimation in future batches.
Note that simple assignment rules may have other advantages; for example, \cite{lei-mhealth} highlights the advantages of simplicity for interpretability in health applications of contextual bandits.

\section{Closing Remarks}
Contextual bandits are poised to play an important role in a wide range of applications:
In these settings, there are many potential sources of bias in estimation of outcome models, not only due to the inherent adaptive data collection, but also due to mismatch between the true data generating process and the outcome model assumptions, and due to prejudice in the training data in form of under-representation or over-representation of certain regions of the context space.	content recommendation in web-services, where the learner wants to personalize recommendations (arm) to the profile of a user (context) to maximize engagement (reward);  online education platforms, where the learner wants to select a teaching method (arm) based on the characteristics of a student (context) in order to maximize the student's scores (reward); and survey experiments, where the learner wants to learn what information or persuasion (arm) influences the responses (reward) of subjects as a function of their demographics, political beliefs, or other characteristics (context).
In these settings, there are many potential sources of bias in estimation of outcome models, not only due to the inherent adaptive data collection, but also due to mismatch between the true data generating process and the outcome model assumptions, and prejudice in the training data in form of under-representation or over-representation of certain regions of the context space in the cold-start training data.
To reduce bias, we proposed new contextual bandit designs which integrate balancing methods from the causal inference literature and parametric and non-parametric model estimation methods with optimism or randomization based exploration methods.
We provided the first regret bound analysis for linear contextual bandits with balancing that matches the theoretical guarantees of the linear contextual bandits with direct model estimation
We showed that contextual bandit designs with randomization based exploration and balancing in model estimation are more robust to the aforementioned sources of bias and have improved learning rates.
We also showed that using simpler, less variable assignment policies, reduces variance in estimation and bias due to confounding and decreases regret.
Through a range of simulations and experiments on real-world datasets, we aimed to highlight key tradeoffs and thus provide methodological guidelines to anyone who wishes to efficiently use contextual bandits for learning personalized policies in practice. 

\section{Acknowledgments}
The authors would like to thank Emma Brunskill for valuable comments on the paper and John Langford, Miroslav Dud{\'\i}k, Akshay Krishnamurthy and Chicheng Zhang for useful discussions regarding the evaluation on classification datasets. 
This research is generously supported by ONR grant N00014-17-1-2131, by the Sloan Foundation, by the ``Arvanitidis in Memory of William K. Linvill'' Stanford Graduate Fellowship in Science \& Engineering and by the Onassis Foundation.

\newpage
\bibliographystyle{plain}
\bibliography{EstimationConsiderationsInContextualBandits}

\newpage
\appendix

\setlength{\parindent}{0pt}
\setlength{\parskip}{6pt}

\vspace*{10pt}
\section{Appendix A: Regret Bound Proofs}

\vspace*{5pt}
\subsection{Auxiliary Results}\label{sec:mg}

We collect in this section all the existing results in the literature for later use.

The first one is the self-normalized bound for vector-valued martingales in~\cite{abbasi2011improved}.
\begin{lemma}\label{lem:self_norm}
	Let $\{F_t\}_{t=0}^{\infty}$ be a filtration. Let $\{\eta_t\}_{t=1}^{\infty}$ be a 
	real-valued stochastic process such that $\eta_t$ is $F_t$ measurable and $\eta_t$ is conditionally R-sub-Guassian for some $R \ge 0$: $\mathbf{E}[e^{\lambda \eta_t} \mid F_{t-1}] \le \exp(\frac{\lambda^2 R^2}{2})$. Let $\{Z_t\}_{t=1}^{\infty}$
	be an $\mathbb{R}^d$-valued stochastic process such that $Z_t$ is $F_{t-1}$ measurable. Assume that $V$ is a $d \times d$ positive definite matrix, and for any integer $t \ge 0$, define:
	$$V_t = V + \sum_{s=1}^t Z_s Z_s^T, S_t = \sum_{s=1}^t \eta_s Z_s.$$
	Then, for any $\delta > 0$, with probability at least $1-\delta$, we have:
	\begin{equation}\label{eq:self_normalized}
	S_t^T V_t^{-1} S_t \le 2R^2 \log(\frac{(\det(V_t))^{\frac{1}{2}} (\det(V))^{\frac{1}{2}}}{\delta}).
	\end{equation}
\end{lemma}

The second one, taken from~\cite{abramowitz1964handbook}, gives large deviation bounds for Guassian random variables.
\begin{lemma}\label{lem:anti_concentration}
	Let $Z$ be a Guassian random variable with mean $m$ and variance $\sigma^2$.
	Then for any real number $r \ge 1$,
	\begin{equation}\label{eqref:guassian_concentration}
	\frac{1}{2\sqrt{\pi}r} e^{-\frac{r^2}{2}} \le \mathbb{P}(|Z -m | > r\sigma) \le
	\frac{1}{\sqrt{\pi}r} e^{-\frac{r^2}{2}}.
	\end{equation}
\end{lemma}

The third one is taken from~\cite{auer2002using}.
\begin{lemma}\label{lem:auer}
	Let $Z^{\prime}$ and $Z$ be two symmetric and positive semidefinite $d \times d$ matrices 
	where the corresponding eigenvalues are $\lambda^{\prime}_1,\dots, \lambda^{\prime}_d$ and  $\lambda_1,\dots, \lambda_d$ respectively.
	If $Z^{\prime} = Z + xx^T$, then the eigenvalues can be arranged in such a way that
	$\lambda_j \le \lambda_j^{\prime}$ and
	$x^T Z^{-1} x \le 10 \sum_{j=1}^d \frac{\lambda_j^{\prime} - \lambda_j}{\lambda_j}$.
\end{lemma}

The next one is a basic fact from matrix algebra (~\cite{horn1990matrix}).
\begin{lemma}\label{lem:matrix}
	Let $M$ be a symmetric, positive definite matrix, and $x, y$ be vectors (all with appropriate dimensions). Then the weighted inner product $\langle \cdot, \cdot \rangle_M$ is defined as $\langle x, y \rangle_M = x^T M y$.
	Furthermore, $\sqrt{\langle x, x \rangle_M}$ is a norm on $x$, which we denote by $\|x\|_M$ and we have
	$|\langle x, y \rangle_M| \le \|x\|_M \|y\|_M$.
\end{lemma}
\subsection{Proof of Theorem 1}

In this section, we provide the proof to the regret bound given in Theorem~1. We start with BLTS, which is more involved.
For ease of exposition, we break the proof into several steps, each of which will be explained.
We start by setting up some notation.
Let $\hat{\theta}_a(t)$ and $\tilde{\theta}_a(t)$ denote the estimated 
mean and the sampled mean for arm $a$ in the BLTS algorithm at time $t$, respectively.
By some algebra, one can show that $(r_a(t) - X_a^T\hat{\theta}_a(t))^T W_a  (r_a(t) - X_a^T\hat{\theta}_a(t)) = \sigma_a^2 \frac{\sum_{i=1}^t w_i}{t} \in [\sigma_a^2 \gamma, \sigma_a^2 \frac{1}{\gamma}]$. Consequently, $B_a(t)$ is only a constant factor of the variance $\mathbf{V}(\hat{\theta}_a(t))$ term and it suffices to focus on $B_a(t)$.
Let $w_t$ denote the thresholded inverse propensity score computed at $t$.
For each arm $a$, an equivalent way of writing the updates in the BLTS algorithm is:
$$\B_a(t+1) = \left\{
\begin{array}{ll}
\B_a(t) + w_t x_{t+1} x_{t+1}^T, & \textbf{if } a \text{ is selected in } t \\
\B_a(t), & \textbf{otherwise},\\
\end{array}
\right.
$$
$$\hat{\theta}_a(t+1) = \left\{
\begin{array}{ll}
\B_a(t+1)^{-1} \sum_{s \in \mathcal{S}_a(t)} w_sx_s r_a(s), & \textbf{if } a \text{ is selected in } t \\
\hat{\theta}_a(t), & \textbf{otherwise},\\
\end{array}
\right.
$$
where $\mathcal{S}_a(t) = \{1 \le s \le t \mid a(s)  = a \}$ keeps track of all the iterations where action $a$ is taken.
Note that an equivalent way of expressing $B_a(t)$ that will be used later is
$B_a(t) = \lambda \mathbf{I}  + \sum_{s \in \mathcal{S}_a(t)} \sqrt{w_s} x_s (\sqrt{w_s}x_s)^T$.
We shall freely use and switch between the update written in this incremental format and the update format given in the main text.

\setcounter{proofstep}{0}

\begin{proofstep}{High concentration bound of estimated means}\label{step:1}
	
	Here we show that $x_t^T \hat{\theta}_a(t)$ concentrates around $x_t^T \theta_a$.
	Furthermore, such concentration is uniform across time $t$ and different arms.
	Specifically, define the event 
	$C = \bigcap_{t=1}^T \bigcap_{a \in \mathcal{A}} C_a(t) $,
	where $C_a(t)$ is the following event:
	$$x_t^T \hat{\theta}_a(t) \in [x_t^T \theta_a - O\Big(  \sqrt{d\log(\frac{KT}{\delta})} \Big)\sqrt{x_t^T (\B_a(t))^{-1} x_t}, x_t^T \theta_a + O\Big(  \sqrt{d\log(\frac{KT}{\delta})} \Big)\sqrt{x_t^T (\B_a(t))^{-1} x_t}].$$
	The event $C_a(t)$ essentially says that $x_t^T \hat{\theta}_a(t)$ is within some constant multiple of standard deviations from $x_t^T \theta_a$ (the true mean reward).
	Note that since $ \tilde{\theta}$ is drawn from a Guassian with mean $\hat{\theta}$
	and variance $\frac{\log\frac{1}{\delta}}{\epsilon} (\B_a(t))^{-1}$,
	the quantity $\sqrt{x_t^T (\B_a(t))^{-1} x_t}$ is the standard deviation (up to some constant multiple) of $x_t^T \tilde{\theta}_a$, which is
	the sample reward associated with arm $a$ estimated by the algorithm BLTS.
	Our goal in this step is to show that the event $C$ occurs with high probability.
	Intuitively, this means that the estimated parameter $\hat{\theta}_a$ is concentrated around the true parameter $\theta_a$: this is an important ingredient for low regret, because if the estimated mean parameters were wrong, then the model maintained by the algorithm would be incorrect, in which case it is impossible to attain any good performance.
	Specifically, we establish that $\mathbf{P} (C) \ge 1 -\frac{\delta}{T}$, 
	$x_t^T \hat{\theta}_a(t) \in [x_t^T \theta_a - O\Big(  \sqrt{d\log(\frac{KT}{\delta})} \Big)\sqrt{x_t^T (\B_a(t))^{-1} x_t},$ $	x_t^T \theta_a + O\Big(  \sqrt{d\log(\frac{KT}{\delta})} \Big)\sqrt{x_t^T (\B_a(t))^{-1} x_t}], \forall t \in \{1,\dots, T\}, \forall a \in \A$.
	To show this, let 
	\begin{eqnarray*}
		\X_s &=& 	\sqrt{w_s} x_s, \quad \eta_s^a =  \sqrt{w_s} (r_a(s) - x_s^T \theta_a)\\
		V_t^a &=& \lambda \mathbf{I}  + \sum_{s \in \mathcal{S}_a(t)} \sqrt{w_s} x_s (\sqrt{w_s}x_s)^T = \lambda \mathbf{I}  + \sum_{s \in \mathcal{S}_a(t)}\X_s \X_s^T\\
		S_t^a &=&  \sum_{s \in \mathcal{S}_a(t)} \sqrt{w_s}x_s \eta_s^a = \sum_{s \in \mathcal{S}_a(t)}  \eta_s^a\X_s.
	\end{eqnarray*}
	Since $r_a(s) - x_t^T \theta_a$ is conditionally sub-Gaussian (conditioned on $\mathcal{F}_{t-1}$), and $w_s$ is bounded, $\sqrt{w_s} (r_a(s) - x_t^T \theta_a)$ is also conditionally sub-Gaussian.
	Let $R^*$ be the universal sub-Gaussian constant for the entire proof.
	It is easy to check that under the canonical filtration $\mathcal{F}_t$,
	all the adaptability assumptions in Lemma~\ref{lem:self_norm} hold:
	specifically, $\X_t$ is adapted to $\mathcal{F}_{t-1}$ and $\eta_t$ is adapted to
	$\mathcal{F}_t$.
	Consequently, by Lemma~\ref{lem:self_norm}, we have for each $a \in \mathcal{A}$,
	with probability at least $1 -\delta$:
	\begin{align}\label{eq:2.2}
	&S_t^T V_t^{-1} S_t \le 2R^2 \log(\frac{(\det(V_t))^{\frac{1}{2}} (\det(\lambda\textbf{I}))^{-\frac{1}{2}}}{\delta}) = 2R^2 \log(\frac{(\det(V_t))^{\frac{1}{2}} (\lambda)^{-\frac{p}{2}}}{\delta})
	\\
	&\le 2R^2 \log(\frac{(\frac{|\mathcal{S}_a(t) |}{\gamma} + \lambda)^{\frac{p}{2}} \lambda^{-\frac{p}{2}}}{\delta}) 
	\le 2R^2 \log(\frac{(\frac{t}{\gamma} + \lambda)^{\frac{p}{2}} \lambda^{-\frac{p}{2}}}{\delta})=
	pR^2 \log(\frac{1 + \frac{t}{\gamma \lambda}}{\delta}),
	\end{align}
	where the last inequality follows because 
	the largest eigenvalue of the matrix $x_t^Tx_t$ can be at most one and hence 
	by a simple calculation, the largest eigenvalue of $V_t^a$ can be at most $\frac{|\mathcal{S}_a(t)|}{\gamma} + \lambda$, which in turn is upper bounded by 
	$\frac{t}{\gamma} + \lambda$. Since the determinant equals the product of all the eigenvalues, and since all the matrices involved are positive-semidefinite, the result follows.
	
	With the above bound in place, we are now ready to bound $x_t^T\hat{\theta}_a(t) - x_t^T \theta_a$. First note that
	$|x_t^T\hat{\theta}_a(t) - x_t^T \theta_a| = |x_t^T (V_t^a)^{-1} (S_t^a - \theta_a)|$.
	Since $V_t^a$ is positive definite, its inverse is positive definite as well.
	Consequently, by Lemma~\ref{lem:matrix}, we have:
	
	\begin{align}
	&|x_t^T\hat{\theta}_a(t) - x_t^T \theta_a| = |x_t^T (V_t^a)^{-1} (S_t^a - \lambda \theta_a)|
	\le \|x_t^T\|_{(V_t^a)^{-1} } \|S_t^a - \lambda \theta_a\|_{(V_t^a)^{-1} } \\
	&=
	\|x_t^T\|_{(B_t^a)^{-1} } \|S_t^a - \lambda \theta_a\|_{(V_t^a)^{-1} } =  \sqrt{x_t^T (\B_a(t))^{-1} x_t} \|S_t^a - \lambda \theta_a\|_{(V_t^a)^{-1} },
	\end{align}
	where the first equality follows from the following algebraic calculation:
	\begin{align}
	&(V_t^a)^{-1} (S_t^a - \theta_a) = (\lambda \mathbf{I}  + \sum_{s \in \mathcal{S}_a(t)}\X_s \X_s^T)^{-1} (S_t^a - \lambda \theta_a) \\
	&= (X_a^T W X_a + \lambda \mathbf{I})(X_a^T W_a r_a-X_a^T W_aX_a\theta_a -\lambda\textbf{I} \theta_a)\\
	& = (X_a^T W X_a + \lambda \mathbf{I})^{-1}(X_a^T W_a r_a)- (X_a^T W X_a + \lambda \mathbf{I})^{-1}(X_a^T W_aX_a\theta_a +\lambda\textbf{I} \theta_a ) = \hat{\theta}_a(t) - \theta_a,
	\end{align}
	and the second inequality follows from Lemma~\ref{lem:matrix}.
	Note that since 
	\begin{align}\label{eq:eigen_value}
	&\|\theta_a\|_{(V_t^a)^{-1}} = \sqrt{ \theta_a^T (V_t^a)^{-1} \theta_a} \le \sqrt{\lambda_{\max} [(V_t^a)^{-1}]} \|\theta_a\| \le \sqrt{\lambda_{\max} [(V_t^a)^{-1}]} \\
	&\le \frac{1}{\sqrt{\lambda_{\min} [V_t^a]}} \le \frac{1}{\sqrt{\lambda_{\min} [\lambda \textbf{I}]}} = \frac{1}{\sqrt{\lambda}},
	\end{align}
	where $\lambda_{\max}$ and $\lambda_{\min}$ denote the maximum eigenvalue and the minimum eigenvalue of a matrix respectively.
	Consequently, we have that with probability at least $1 -\delta$:
	\begin{align}
	&|x_t^T\hat{\theta}_a(t) - x_t^T \theta_a| \le  \sqrt{x_t^T (\B_a(t))^{-1} x_t} \|S_t^a - \lambda \theta_a\|_{(V_t^a)^{-1} } \\
	&\le  \sqrt{x_t^T (\B_a(t))^{-1} x_t} \Big(\|S_t^a\|_{(V_t^a)^{-1} } + \lambda \|\theta_a\|_{(V_t^a)^{-1} }\Big)\\
	&  \le \sqrt{x_t^T (\B_a(t))^{-1} x_t} \Big(\|S_t^a\|_{(V_t^a)^{-1} } + \lambda \frac{1}{\sqrt{\lambda} }\Big) \le \sqrt{x_t^T (\B_a(t))^{-1} x_t} \Big( \sqrt{d}R \sqrt{\log(\frac{1 + \frac{t}{\gamma \lambda}}{\delta})}  + \sqrt{\lambda} \Big)\\
	& \le \sqrt{x_t^T (\B_a(t))^{-1} x_t} \Big( \sqrt{d}R \sqrt{\log(\frac{1 + \frac{T}{\gamma \lambda}}{\delta})}  + \sqrt{\lambda} \Big)\\
	\end{align}
	where the second inequality follows from the triangle inequality of a norm, the third inequality follows from Equation~\eqref{eq:eigen_value} and the last inequality follows from Equation~\eqref{eq:2.2}.
	Now take $\delta$ to be $\frac{\delta}{KT^2}$, and absorbing all the constants into the big-$O$, we have, with probability at least $1 -\frac{\delta}{KT^2}$,
	$$\mathbf{P}(C_a(t)) \le  \sqrt{x_t^T (\B_a(t))^{-1} x_t} \Big( \sqrt{d}R \sqrt{\log(\frac{KT^2 + K\frac{T^3}{\gamma \lambda}}{\delta})}  + \sqrt{\lambda} \Big) = \sqrt{x_t^T (\B_a(t))^{-1} x_t} O\Big(  \sqrt{d\log(\frac{KT}{\delta})} \Big).$$ 
	By a union bound, we have:
	$\p(C) = 1 -\mathbf{P}(\bigcup_{t=1}^T \bigcup_{a \in \mathcal{A}} C_a^c(t)) \le \sum_{t=1}^T\sum_{a \in \mathcal{A}} \mathbf{P}(C_a^c(t)) \le \frac{\delta}{T}$.
	Consequently, we have with probability at least $1 - \frac{\delta}{T}$,
	$x_t^T \hat{\theta}_a(t) \in [x_t^T \theta_a - O\Big(  \sqrt{d\log(\frac{KT}{\delta})} \Big)\sqrt{x_t^T (\B_a(t))^{-1} x_t},$ $	x_t^T \theta_a + O\Big(  \sqrt{d\log(\frac{KT}{\delta})} \Big)\sqrt{x_t^T (\B_a(t))^{-1} x_t}], \forall t \in \{1,\dots, T\}, \forall a \in \A$.
\end{proofstep}

\begin{proofstep}{High concentration bound of sampled means}\label{step:2}
	
	The above step establishes that $\hat{\theta}_a$ concentrates around the true $\theta_a$. 	
	We now establish that $\tilde{\theta}_a$ is concentrates around $\hat{\theta}_a$.
	This concentration is very much to be expected, because $\tilde{\theta}_a$ is exactly drawn from a Guassian with mean $\hat{\theta}_a$. Consequently, since Guassian random variables concentrate around its mean (Lemma~\ref{lem:anti_concentration}), by choosing an appropriate multiple of its standard deviation, we can get the desired concentration probability relatively easily.
	Specifically, here we have that with probablity at least $1 -\frac{1}{T}$, for all $t = 1,\dots, T$, and for all $a \in \A$:
	\begin{align}
	x_t^T \tilde{\theta}_a(t) \in &[x_t^T \hat{\theta}_a - O\Big(  \sqrt{d \frac{\log\frac{1}{\delta}}{\epsilon} \log(dKT)}) \Big)\sqrt{x_t^T (\B_a(t))^{-1} x_t}, \\
	&x_t^T \hat{\theta}_a + O\Big(  \sqrt{d \frac{\log\frac{1}{\delta}}{\epsilon} \log(dKT)}) \Big)\sqrt{x_t^T (\B_a(t))^{-1} x_t}].
	\end{align}
	
	To see this, denote $E_a(t)$ to be the above event. Then by a straightfoward application of Lemma~\ref{lem:anti_concentration}, one can easily check that
	$$\p\Big(\|B_a(t)^{0.5}\Big(\tilde{\theta}_a(t) - \hat{\theta}_a(t) \Big)\| \ge \sqrt{\frac{\log\frac{1}{\delta}}{\epsilon}} \sqrt{4d\log(dNT)}   \Big) \le \frac{1}{KT^2}.$$
	Consequently, by a further union bound across all $a$ and across all $t$, 
	we have with probability at least $1 -\frac{1}{T}$, 
	$\|B_a(t)^{0.5}\Big(\tilde{\theta}_a(t) - \hat{\theta}_a(t) \Big)\| \le \sqrt{\frac{\log\frac{1}{\delta}}{\epsilon}} \sqrt{4d\log(dKT)} $.
	Note that when this holds, we can easily bound $|x_t^T \tilde{\theta}_a(t) - x_t^T \hat{\theta}_a|$ as follows:
	\begin{align}
	& |x_t^T \tilde{\theta}_a(t) - x_t^T \hat{\theta}_a| =
	|x_t^T B_a(t)^{-0.5} B_a(t)^{0.5} (\tilde{\theta}_a(t) - \hat{\theta}_a)| \\
	& \le \|x_t^T B_a(t)^{-0.5}\| \|B_a(t)^{0.5} (\tilde{\theta}_a(t) - \hat{\theta}_a)\| 
	\le \sqrt{x_t^T B_a(t)^{-1} x_t} \sqrt{\frac{\log\frac{1}{\delta}}{\epsilon}} \sqrt{4d\log(dKT)}.
	\end{align}
	Putting the above two pieces together and using $O(\cdot)$ to simplify all the constants, yields that with probablity at least $1-\frac{1}{T}$,
	$x_t^T \tilde{\theta}_a(t) \in [x_t^T \hat{\theta}_a - O\Big(  \sqrt{d \frac{\log\frac{1}{\delta}}{\epsilon} \log(dKT)}) \Big)\sqrt{x_t^T (\B_a(t))^{-1} x_t}, 
	x_t^T \hat{\theta}_a + O\Big(  \sqrt{d \frac{\log\frac{1}{\delta}}{\epsilon} \log(dKT)}) \Big)\sqrt{x_t^T (\B_a(t))^{-1} x_t}].$
\end{proofstep}

\begin{proofstep}{Bounding regret in terms of standard deviations}
	
	The previous two steps combined together establish that the samples $\tilde{\theta}_a(t)$ produced by BLTS are close to
	the true $\theta_a$. An immediate consequence of this is that we can then bound the instantaneous regret $ir(t)$ at time $t$ in terms of the standard deviations. 
	More specifically, under the above two concentration events, for each arm $a$,
	$x_t^T \tilde{\theta}_a(t)$ differs from $x_t^T \hat{\theta}_a$ by at most $O\Big(  \sqrt{d \frac{\log\frac{1}{\delta}}{\epsilon} \log(dKT)}) \Big)\sqrt{x_t^T (\B_a(t))^{-1} x_t}$
	and $x_t^T \hat{\theta}_a(t)$ differs from $x_t^T \theta_a$ by at most $O\Big(  \sqrt{d\log(\frac{KT}{\delta})} \Big)\sqrt{x_t^T (\B_a(t))^{-1} x_t}]$. Consequently, $x_t^T \tilde{\theta}_a(t)$ differs from $x_t^T \theta_a$
	by at most $O\Big(  \sqrt{d \frac{\log\frac{1}{\delta}}{\epsilon} \log(dKT)}) +  \sqrt{d\log(\frac{KT}{\delta})}\Big)\sqrt{x_t^T (\B_a(t))^{-1} x_t}$.
	Next, since action $a$ is chosen at time $t$, it must be that $x^T_t\tilde{\theta}_a(t)$ yields the largest value, and in particualr,  $x^T_t\tilde{\theta}_a(t) \ge x^T_t \theta_{a^*(t)}$.
	Putting the above discussion together, we can bound the instantaneous regret as follows:
	\begin{align}
	&ir(t) = x_t^T(\theta_{a^*(t)} - \theta_{a(t)}) \\
	&\le  O\Big(  \sqrt{d \frac{\log\frac{1}{\delta}}{\epsilon} \log(dKT)}) +  \sqrt{d\log(\frac{KT}{\delta})}\Big)\Big(\sqrt{x_t^T (\B_{a(t)}(t))^{-1} x_t} +  \sqrt{x_t^T (\B_{a^*(t)}(t))^{-1} x_t} \Big),
	\end{align}
	where $a(t)$ is the arm chosen at $t$ and $a^*(t)$ is the optimal arm at $t$.
	Consequently, 
	\begin{align}
	R(T) = \sum_{t=1}^T ir(t) &\le O\Big(  \sqrt{d \frac{\log\frac{1}{\delta}}{\epsilon} \log(dKT)}) +  \sqrt{d\log(\frac{KT}{\delta})}\Big)\sum_{t=1}^T\Big(\sqrt{x_t^T (\B_{a(t)}(t))^{-1} x_t} +  \sqrt{x_t^T (\B_{a^*(t)}(t))^{-1} x_t} \Big)\\
	& = O\Big(  \sqrt{d \frac{\log\frac{1}{\delta}}{\epsilon} \log(dKT)}) \Big)\sum_{t=1}^T\Big(\sqrt{x_t^T (\B_{a(t)}(t))^{-1} x_t} +  \sqrt{x_t^T (\B_{a^*(t)}(t))^{-1} x_t} \Big) \\
	&= \tilde{O}\Big(  \sqrt{ \frac{d}{\epsilon} } \Big)\sum_{t=1}^T\Big(\sqrt{x_t^T (\B_{a(t)}(t))^{-1} x_t} +  \sqrt{x_t^T (\B_{a^*(t)}(t))^{-1} x_t} \Big).
	\end{align}
	
	The rest of the proof can then be completed by bounding the sum in the right-hand side of the above equation.
	First, following a similar analysis (with differences only in constants) as in~\cite{agrawal-lints}, one can use a martingale based approach to bound
	$\sqrt{x_t^T (\B_{a^*(t)}(t))^{-1} x_t} $ in terms of $\sqrt{x_t^T (\B_{a(t)}(t))^{-1} x_t}$.
	This is done by dividing arms into two different categories and do a careful analysis of the bound in each case.
	The final bound on the sum, after dropping all the lower order terms is that
	\begin{equation}\label{eq:help}
	\sum_{t=1}^T\Big(\sqrt{x_t^T (\B_{a(t)}(t))^{-1} x_t} +  \sqrt{x_t^T (\B_{a^*(t)}(t))^{-1} x_t} \Big)\le \sum_{t=1}^T (1 + O(\sqrt{T^{\epsilon}}))\sqrt{x_t^T (\B_{a(t)}(t))^{-1} x_t}.
	\end{equation}
	
	Next, in~\cite{chu2011contextual}, it is shown that for each arm $a$, define $\mathcal{T}_a = \{1 \le t \le T \mid a(t) =a\}$, then the following holds:
	$$ \sum_{t\in \mathcal{T}_a} \sqrt{x_t^T (\B_{a(t)}(t))^{-1} x_t}\le 5 \sqrt{d N_a(T) \log (N_a(T))},$$
	where $N_a(T) = | \mathcal{T}_a | $ is the total number of times arm $a$ is selected (note in particular $\sum_{a \in \A} N_a(T) = T$).
	Consequently, summing over all $a$, we have:
	\begin{equation}\label{eq:useful}
	\sum_{t=1}^T \sqrt{x_t^T (\B_{a(t)}(t))^{-1} x_t} = \sum_{a \in \A} \sum_{t\in \mathcal{T}_a} \sqrt{x_t^T (\B_{a(t)}(t))^{-1} x_t} \le \sum_{a \in \A} 5 \sqrt{d N_a(T) \log (N_a(T))} = O(\sqrt{dKT\log T}).
	\end{equation}
	Consequently, combining the above inequality with Equation~\eqref{eq:help}, we have:
	\begin{align}
	&\sum_{t=1}^T\Big(\sqrt{x_t^T (\B_{a(t)}(t))^{-1} x_t} +  \sqrt{x_t^T (\B_{a^*(t)}(t))^{-1} x_t} \Big)\le  (1 + O(\sqrt{T^{\epsilon}}))O(\sqrt{pNT\log T}) \\
	& = O(\sqrt{dKT^{1+\epsilon}\log T}) =\tilde{O}(\sqrt{dKT^{1+\epsilon}}).
	\end{align}
	Finally, combining the preceding inequality with Equation (2.23), we obtain the final regret bound (note that all the inequalities hold with probability $1 - 2\delta$, since we need the concentration events to hold true):
	$$R(T) = \tilde{O}\Big(  \sqrt{ \frac{d}{\epsilon} } \Big)\tilde{O}(\sqrt{dKT^{1+\epsilon}}) = \tilde{O}(d\frac{\sqrt{KT^{1+\epsilon}}}{\epsilon}).$$
\end{proofstep}	

Next, the proof of regret bound for BLUCB follows closely to that of LinUCB in~\cite{chu2011contextual} and LinRel in~\cite{auer2002using},
where one divides the BLUCB into two parts for analysis, the first part is BaseBLUCB (which corresponds to BaseLinUCB) and the second part is SuperBLUCB (which is the same as SuperLinUCB in~\cite{chu2011contextual}). Hence, the analysis only needs to be adjusted for BaseBLUCB (and we omit the discussion of SuperBLUCB as it is the same as SuperLinUCB in~\cite{chu2011contextual}).
We start by noting that parameters in BLUCB follow the same update as BLST, which are
written as follows:
$$\B_a(t+1) = \left\{
\begin{array}{ll}
\B_a(t) + w_t x_{t+1} x_{t+1}^T, & \textbf{if } a \text{ is selected in } t \\
\B_a(t), & \textbf{otherwise},\\
\end{array}
\right.
$$
$$\hat{\theta}_a(t+1) = \left\{
\begin{array}{ll}
\B_a(t+1)^{-1} \sum_{s \in \mathcal{S}_a(t)} w_sx_s r_a(s), & \textbf{if } a \text{ is selected in } t \\
\hat{\theta}_a(t), & \textbf{otherwise},\\
\end{array}
\right.$$
where $\mathcal{S}_a(t) = \{1 \le s \le t \mid a(s)  = a \}$ keeps track of all the iterations where action $a$ is taken.

Consequently, from Step 1 above, and with $\alpha = \sqrt{\log \frac{TK}{\delta}}$, we know that with probability at least $1 - \frac{\delta}{T}$, for each $a \in \mathcal{A}$ and each $t = 1, 2, \dots, T$:
$$|x_t^T \hat{\theta}_a(t)  - x_t^T \theta_a| \le  O\Big(  \sqrt{\log(\frac{KT}{\delta})} \Big)\sqrt{x_t^T (\B_a(t))^{-1} x_t}.$$
Furthermore, by Equation~\ref{eq:useful} in Step 3, we have:
\begin{equation}
\sum_{t=\in \Phi_{T+1}} \sqrt{x_t^T (\B_{a(t)}(t))^{-1} x_t}  = O(\sqrt{dK|\Phi_{T+1}|\log |\Phi_{T+1}|}).
\end{equation}
With these two main ingredients in place, the rest of the proof follows the same steps 
as in~\cite{chu2011contextual} (which in turn follows~\cite{auer2002using}), giving a $O(\sqrt{TdK \log^3 (\frac{\log KT\log(T)}{\delta}})$ regret bound.

\begin{algorithm}[t]
	\caption{BaseBLUCB at Step $t$} 
	\label{alg:BLUCBB}
	\begin{algorithmic}[1]
		\State{Inputs:}  $\alpha > 0$ and $\Phi_t \subset \{1,2, \dots, t-1\}$
		\For {$\tau = 1, \dots, t-1$}
		\State Estimate $\hat{p}_a(x_\tau)$ and set $w_\tau = \frac{1}{\max(\gamma, \hat{p}_a(x_\tau))}$.
		\EndFor
		
		\State 
		$B_a(t) \leftarrow \lambda \mathbf{I}  + \sum_{s \in \mathcal{S}_a(t)} \mathbf{1}_{s \in \Phi_t}\sqrt{w_s} x_s (\sqrt{w_s}x_s)^T$, for each $a$.
		\State $\hat{\theta}_a(t) \leftarrow B_a(t)^{-1} \sum_{s \in \mathcal{S}_a(t)} \mathbf{1}_{s \in \Phi_t}w_sx_s r_a(s)$, for each $a$.

		\State Observe feature $x_t$.
		\State $s_a(t) \leftarrow \alpha \sqrt{x_t^T B_a^{-1}(t) x_t}$,
		for each $a$.
		\State $r_a(t) \rightarrow \hat{\theta}_a^T x_t$, for each $a$.
		
	\end{algorithmic}
\end{algorithm}
\newpage
\section{Experiments on Multi-Class Classification Datasets}
\vspace*{10pt}

\sloppy

We use 300 multiclass datasets from the Open Media Library (OpenML). The full list of datasets in alphabetical order is:

\vspace*{5pt}
{
\small
\noindent 
2dplanes,
abalone (183),
abalone (720),
acute-inflammations,
Agrawal1,
aids,
ailerons,
airlines,
analcatdata\_apnea2,
analcatdata\_apnea3,
analcatdata\_asbestos,
analcatdata\_authorship,
analcatdata\_challenger,
analcatdata\_creditscore,
analcatdata\_dmft,
analcatdata\_germangss,
analcatdata\_japansolvent,
analcatdata\_michiganacc,
analcatdata\_olympic2000,
analcatdata\_seropositive,
analcatdata\_vehicle,
analcatdata\_vineyard,
analcatdata\_wildcat,
AP\_Breast\_Kidney,
AP\_Breast\_Omentum,
AP\_Breast\_Ovary,
AP\_Colon\_Kidney,
AP\_Colon\_Lung,
AP\_Endometrium\_Breast,
AP\_Endometrium\_Kidney,
AP\_Endometrium\_Ovary,
AP\_Endometrium\_Prostate,
AP\_Lung\_Kidney,
AP\_Lung\_Uterus,
AP\_Omentum\_Lung,
AP\_Omentum\_Prostate,
AP\_Omentum\_Uterus,
AP\_Ovary\_Kidney,
AP\_Ovary\_Lung,
AP\_Prostate\_Ovary,
AP\_Uterus\_Kidney,
ar3,
ar4,
ar5,
ar6,
arsenic-male-bladder,
arsenic-male-lung,
artificial-characters,
Australian,
autoPrice,
balance-scale,
banana,
baskball,
bodyfat,
bolts,
boston (853),
boston (872),
boston\_corrected,
BurkittLymphoma,
cal\_housing,
car,
chatfield\_4,
chscase\_adopt,
chscase\_census2,
chscase\_census3,
chscase\_vine2,
cmc,
codrna,
codrnaNorm,
collins (478),
collins (987),
confidence (468),
confidence (1015),
covertype (180),
covertype (293),
cpu,
cpu\_small,
delta\_ailerons,
delta\_elevators,
desharnais,
diabetes,
diabetes\_numeric,
diggle\_table\_a2,
disclosure\_x\_bias,
disclosure\_x\_noise,
dresses-sales,
eating,
ecoli,
electricity,
elevators,
elusage,
energy-efficiency,
first-order-theorem-proving,
fl2000,
flags (285),
flags (1012),
fri\_c0\_1000\_25,
fri\_c0\_1000\_5,
fri\_c0\_100\_10,
fri\_c0\_100\_50,
fri\_c0\_250\_10,
fri\_c0\_250\_25,
fri\_c0\_250\_5,
fri\_c0\_500\_25,
fri\_c0\_500\_50,
fri\_c1\_1000\_5,
fri\_c1\_1000\_50,
fri\_c1\_100\_10,
fri\_c1\_100\_5,
fri\_c1\_100\_50,
fri\_c1\_250\_10,
fri\_c1\_250\_5,
fri\_c1\_250\_50,
fri\_c1\_500\_10,
fri\_c1\_500\_5,
fri\_c1\_500\_50,
fri\_c2\_1000\_10,
fri\_c2\_1000\_25,
fri\_c2\_1000\_5,
fri\_c2\_1000\_50,
fri\_c2\_100\_10,
fri\_c2\_100\_25,
fri\_c2\_100\_5,
fri\_c2\_100\_50,
fri\_c2\_250\_10,
fri\_c2\_250\_25,
fri\_c2\_250\_5,
fri\_c2\_250\_50,
fri\_c2\_500\_10,
fri\_c2\_500\_25,
fri\_c2\_500\_5,
fri\_c3\_1000\_10,
fri\_c3\_1000\_25,
fri\_c3\_1000\_5,
fri\_c3\_1000\_50,
fri\_c3\_100\_10,
fri\_c3\_100\_5,
fri\_c3\_250\_10,
fri\_c3\_250\_5,
fri\_c3\_500\_10,
fri\_c3\_500\_25,
fri\_c3\_500\_5,
fri\_c4\_1000\_10,
fri\_c4\_1000\_50,
fri\_c4\_100\_10,
fri\_c4\_100\_100,
fri\_c4\_100\_50,
fri\_c4\_250\_25,
fri\_c4\_500\_10,
fri\_c4\_500\_100,
fri\_c4\_500\_50,
gina\_agnostic,
gina\_prior2,
glass,
grub-damage (338),
grub-damage (1026),
hayes-roth,
heart-statlog,
houses,
humandevel,
hutsof99\_child\_witness,
hutsof99\_logis,
Hyperplane\_10\_1E-4,
iris,
JapaneseVowels,
jEdit\_4.0\_4.2,
jEdit\_4.2\_4.3,
kc1,
kc1-binary,
kc2,
kin8nm,
kr-vs-k,
kr-vs-kp,
kropt,
leaf,
letter (6),
letter (977),
leukemia,
lowbwt,
lupus,
machine\_cpu,
MagicTelescope,
mammography,
mc2,
meta\_all.arff,
meta\_ensembles.arff,
mfeat-factors,
mfeat-fourier,
mfeat-karhunen (16),
mfeat-karhunen (1020),
mfeat-morphological (18),
mfeat-morphological (962),
mfeat-pixel (20),
mfeat-pixel (1022),
mfeat-zernike (22),
mfeat-zernike (995),
monks-problems-3,
mu284,
musk,
mv,
mw1,
MyIris,
newton\_hema,
no2,
nursery (26),
nursery (959),
optdigits,
OVA\_Endometrium,
OVA\_Ovary,
OVA\_Prostate,
page-blocks,
pasture (339),
pasture (964),
pc1,
pc2,
pc3,
pc4,
pendigits,
PieChart1,
PieChart3,
PieChart4,
PizzaCutter3,
plasma\_retinol,
pm10,
pollen,
pollution,
prnn\_cushings,
prnn\_fglass (952),
prnn\_fglass (996),
puma32H,
puma8NH,
pwLinear,
pyrim,
quake (209),
quake (772),
quake (948),
rabe\_131,
rabe\_148,
rabe\_265,
rabe\_266,
rabe\_97,
rmftsa\_ladata,
rmftsa\_sleepdata (679),
rmftsa\_sleepdata (741),
rsctc2010\_1,
rsctc2010\_2,
satellite\_image,
satimage,
scene,
schlvote,
SEA(50),
SEA(50000),
segment,
sensory,
sleuth\_case1102,
sleuth\_case1201,
sleuth\_case1202,
sleuth\_case2002,
sleuth\_ex1605,
sleuth\_ex2016 (682),
sleuth\_ex2016 (862),
socmob,
sonar,
space\_ga,
spambase,
spectrometer (313),
spectrometer (754),
Stagger1,
Stagger2,
Stagger3,
stock,
strikes,
sylva\_agnostic,
sylva\_prior,
synthetic\_control (377),
synthetic\_control (1004),
tae,
teachingAssistant,
tecator,
tic-tac-toe,
tr21.wc,
tr23.wc,
vehicle (54),
vehicle (994),
vehicle\_sensIT,
vehicleNorm,
vinnie,
visualizing\_environmental,
visualizing\_galaxy,
visualizing\_hamster,
visualizing\_soil,
vowel,
waveform-5000,
white-clover,
wind,
wind\_correlations,
wine,
wine\_quality,
witmer\_census\_1980,
zoo}

\vspace*{5pt}
The datasets vary in number of observations, number of classes and number of features. Table \ref{appstats} summarizes the characteristics of these benchmark datasets. 
\begin{table}[H]
	\centering
	\begin{tabular}{|c|c|}
		\hline
		Observations & Datasets \\
		\hline 
		$\leq 100$ & 58 \\
		\hline
		$> 100$ and $\leq 1000$  & 152  \\ 
		\hline
		$> 1000$ and $\leq 10000$ & 57  \\ 
		\hline
		$> 10000$ & 33 \\
		\hline
	\end{tabular}
	\\
	\vspace*{10pt}
	\begin{tabular}{|c|c|}
		\hline
		Classes & Count \\
		\hline 	
		$2$ & 243 \\
		\hline
		$> 2 \text{ and } 10$  & 48  \\ 
		\hline
		$ > 10 $ & 9  \\
		\hline
	\end{tabular}
	\qquad
	\begin{tabular}{|c|c|}
		\hline
		Features & Count \\
		\hline 	
		$\leq 10$ & 154 \\
		\hline
		$> 10 \text{ and } \leq 100$  & 106  \\ 
		\hline
		$> 100$  & 40  \\
		\hline
	\end{tabular}
	\caption{Characteristics of the 300 datasets used for the multiclass classification with bandit feedback experiment.}
	\label{appstats}
\end{table}

In the main paper, we focused on the family of contextual bandits with linear realizability assumption and on how to improve model estimation in linear contextual bandits using balancing. 
Hence, we compared our algorithms, balanced linear Thompson sampling (BLTS) and balanced linear UCB (BLUCB), with LinTS \cite{agrawal-lints} and LinUCB \cite{li-linucb}, which are baselines that belong in the family of contextual bandits with linear realizability assumption and have strong theoretical guarantees.
Here, apart from BLTS, BLUCB, LinTS and LinUCB, we also evaluate the policy-based ILOVETOCONBANDITS (ILTCB) from \cite{agarwal-ilovetoconbandits} that does not estimate a model, but instead it assumes access to an oracle for solving fully supervised cost-sensitive classification problems and achieves the statistically optimal regret guarantee.

\begin{figure}[H]
	\centering
	\includegraphics[width=1\linewidth]{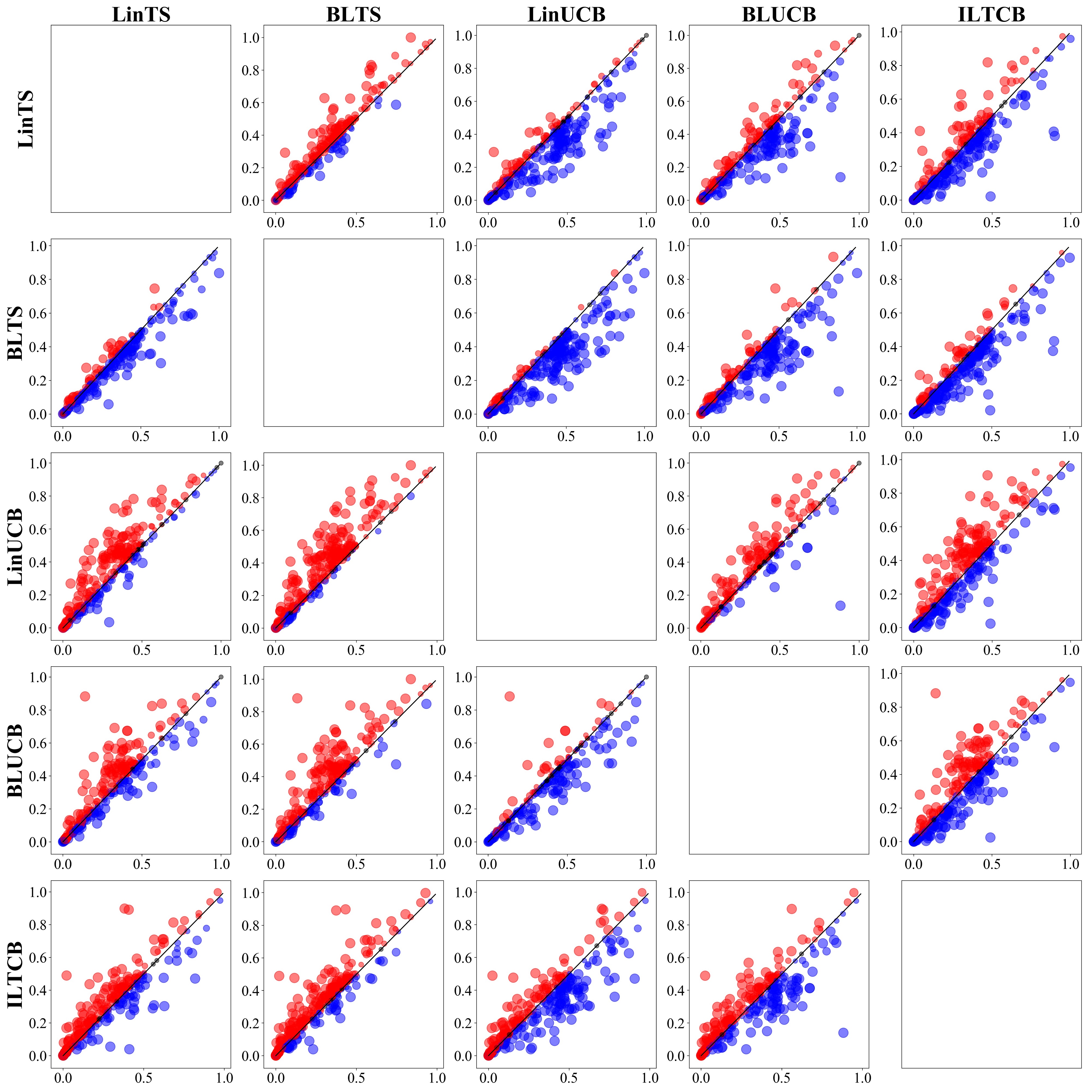}
	\caption{Pairwise comparison of LinTS, BLTS, LinUCB, BLUCB and ILTCB on the 300 classification datasets. BLTS outperforms LinTS, LinUCB, BLUCB and ILTCB.}
	\label{appfig:all_vs_all}
\end{figure}

Figure \ref{appfig:all_vs_all} shows the pairwise comparison of LinTS, BLTS, LinUCB, BLUCB and ILTCB on the 300 classification datasets. Each point corresponds to a dataset. The $x$ coordinate is the normalized cumulative regret of the column bandit and the $y$ coordinate is the normalized cumulative regret of the row bandit. The point is blue when the row bandit has smaller normalized cumulative regret and wins over the column bandit indicated. The point is red when the row bandit loses from the column bandit. The point's size grows with the significance of the win or loss. Table \ref{apptab} presents the pairwise comparison of LinTS, BLTS, LinUCB, BLUCB and ILTCB on the 300 classification datasets in table-form.

In LinTS, LinUCB, BLTS and BLUCB, the ridge regularization parameter $\lambda$ is chosen via cross-validation every time the model is updated. 
In LinTS and BLTS, the constant $\alpha$ is optimized among values $0.25, 0.5, 1$, while in LinUCB and BLUCB  the constant $\alpha$ is optimized among values $1, 2, 4$. 
In BLTS and BLUCB, the propensity threshold $\gamma$ is optimized among the values $0.01, 0.05, 0.1, 0.2$.
In ILTCB, the parameter $\mu$ of algorithm 1 in \cite{agarwal-ilovetoconbandits} is optimized among the values $0.01, 0.1, 1$. 

{
\begin{table}[H]
	\footnotesize
	\centering
	\begin{tabular}{|c|c|c|c|c|c|}
		\hline
		(down vs. right) & LinTS & BLTS & LinUCB & BLUCB & ILTCB \\
		\hline 
		LinTS & W=0, L=0 & W=58, L=233 & W=180, L=96 & W=149, L=130 & W=191, L=95 \\
		\hline 
		BLTS & W=233, L=58 & W=0, L=0 & W=232, L=55 & W=191, L=92 & W=222, L=60 \\
		\hline 
		LinUCB & W=96, L=180 & W=55, L=232 & W=0, L=0 & W=56, L=185 & W=135, L=153  \\
		\hline 
		BLUCB & W=130, L=149 & W=92, L=191 & W=185, L=56 & W=0, L=0  & W=153, L=129 \\
		\hline
		ILTCB & W=95, L=191 & W=60, L=222 & W=153, L=135 & W=129, L=153 & W=0, L=0 \\
		\hline
	\end{tabular}
	\caption{Number of the 300 classification datasets in which the contextual bandit algorithm of the row name wins over (W) or loses from (L)  the contextual bandit algorithm of the column name. The $\text{300}-\text{W}-\text{L}$ remaining datasets are ties.}
	\label{apptab}
\end{table}
}

\newpage
\section{Bayesian LASSO Contextual Bandit}
\vspace*{10pt}
We now provide a Bayesian way of using LASSO estimation in a Thompson sampling contextual bandit inspired by \cite{park2008bayesian}. The theoretical analysis of Bayesian LASSO contextual bandit may be proven more straightforward than the theoretical analysis of bootstrap LASSO contextual bandit, though we leave this analysis for future work.

We model the reward as a linear function of the context
$r = x^T \theta + \epsilon, \epsilon \sim \N(0, \sigma^2)$, 
where $\theta$ is $p$-dimensional and sparse.
LASSO estimates can be interpreted as posterior mode estimates when the coefficients have \textbf{iid} Laplace priors \cite{tibshirani-lasso}.
\cite{park2008bayesian} propose an expanded hierarchy with
conjugate normal priors for the regression parameters and independent exponential priors on their variances to perform Gibbs sampling on this posterior.
Algorithm \ref{bayesianLASSO} proposes a contextual bandit that uses the Gibbs sampler hierarchy of \cite{park2008bayesian}.

\begin{algorithm}[h]
	\caption{Bayesian LASSO Thompson sampling} 
	\label{bayesianLASSO}
	\begin{algorithmic}[1]
		\State \textbf{Input:} Regularization parameter $\lambda > 0$ 
		\State Set $\textbf{X}_a \leftarrow$ empty matrix, $\textbf{r}_a \leftarrow$ empty vector $\forall a \in \mathcal{A}$
		\State Sample \textbf{iid} $\tau^2_{a,1}, \dots, \tau^2_{a,p} \sim \prod_{j=1}^{p} \frac{\lambda^2}{2} e^{-\frac{\lambda^2 \tau_j^2}{2}} d\tau_j^2$ $\forall a \in \mathcal{A}$
		\State Set $\textbf{D}_{\tau_{a}} \leftarrow \text{diag}(\tau^2_{a,1}, \dots, \tau^2_{a,p})$ $\forall a \in \mathcal{A}$
		\State Sample $\theta_{a, 1} \sim \N(0, \sigma^2 \textbf{D}_{\tau_{a}})$
		\For {$t = 1, 2, \dots, T$}
		\State Observe $x_t$
		\State Select $a \leftarrow \arg\max_{a\in\A} x_t^T \theta_{a, t}$
		\State Observe reward $r_t(a)$.
		\State $\textbf{X}_a \leftarrow [\textbf{X}_a : x_t^T]$
		\State $\textbf{r}_a \leftarrow [\textbf{r}_a : r_t(a)]$ 
		\For {$k = 1, \dots, K$ Gibbs sampling iterations}
		\If {$k = 1$} 
		\State $\theta_{a, t+1}^k \leftarrow \theta_{a, t}$
		\EndIf
		\State Sample $\left(\tau^k_{a, j}\right)^2 \sim \text{InverseGaussian}\left(\mu' = \sqrt{\frac{\lambda^2 \sigma^2}{\left(\theta^k_{a, t+1, j}\right)^2}}, \lambda' = \lambda^2\right)$ $\forall j = 1, \dots, p$
		\State Set $\textbf{D}_{\tau^k_{a}} \leftarrow \text{diag}\left(\left(\tau^k_{a,1}\right)^2, \dots, \left(\tau^k_{a,p}\right)^2\right)$ and $\textbf{A}_a = \textbf{X}_a^T \textbf{X}_a + \textbf{D}_{\tau^k_{a}}$
		\State Sample $\theta_{a, t+1}^k \sim \N(\textbf{A}_a^{-1}\textbf{X}_a^T \textbf{r}_a, \sigma^2 \textbf{A}_a^{-1})$ 
		\EndFor
		\State Sample $\theta_{a, t+1} \sim \left(\theta^1_{a, t+1}, \dots, \theta^K_{a, t+1}\right)$.
		\State Set $\theta_{a', t+1} \leftarrow \theta_{a', t}$ $\forall a' \in \A \backslash a$
		\EndFor
	\end{algorithmic}
\end{algorithm}

\end{document}